\documentclass[11pt]{article}

\usepackage[margin=1in]{geometry}
\usepackage{amsmath,amssymb,amsthm,mathtools}
\usepackage{booktabs}
\usepackage{enumitem}
\usepackage{microtype}
\usepackage{flafter}
\usepackage{xcolor}
\usepackage{tikz}
\usetikzlibrary{decorations.pathreplacing,arrows.meta,positioning}
\usepackage{graphicx}
\usepackage[colorlinks=true,linkcolor=blue!60!black,citecolor=blue!60!black,urlcolor=blue!60!black]{hyperref}

\newtheorem{theorem}{Theorem}[section]
\newtheorem{lemma}[theorem]{Lemma}

\newtheorem{claim}[theorem]{Claim}
\newtheorem{fact}[theorem]{Fact}
\theoremstyle{definition}
\newtheorem{definition}[theorem]{Definition}
\theoremstyle{remark}
\newtheorem{remark}[theorem]{Remark}

\newcommand{\cC}{\mathcal{C}}
\newcommand{\cF}{\mathcal{F}}
\newcommand{\cX}{\mathcal{X}}
\newcommand{\cH}{\mathcal{H}}
\newcommand{\cP}{\mathcal{P}}
\newcommand{\cK}{\mathcal{K}}
\newcommand{\cM}{\mathcal{M}}
\newcommand{\cS}{\mathcal{S}}
\newcommand{\Pool}{\mathrm{Pool}}
\newcommand{\VC}{\mathrm{VC}}
\newcommand{\err}{\mathrm{err}}
\newcommand{\set}{\mathrm{set}}
\newcommand{\ind}{\mathbf{1}}
\newcommand{\wt}{\widetilde}

\title{\texorpdfstring{\textbf{Near-Quadratic Sample Compression for All VC Classes}\thanks{This result was obtained autonomously by the Cogentic multi-agent orchestration~\cite{CGJLMVW26} running on Gemini~4 Carbon; see the \hyperref[sec:ai-disclosure]{AI Disclosure} at the end of the paper for a detailed discussion.}}{Near-Quadratic Sample Compression for All VC Classes}}
\author{
  Yang Cai\thanks{Also affiliated with Yale University.} \quad
  Vineet Gupta \quad
  Yanchen Jiang \quad
  Christopher Liaw \quad
  Aranyak Mehta \\[0.3em]
  Shay Moran\thanks{Also affiliated with the Technion.} \quad
  Grigoris Velegkas \quad
  Di Wang \quad
  Manfred K. Warmuth \\[0.45em]
  {\normalsize Google}
}
\date{}

\begin{document}
\maketitle
\vspace{-2.2em}

\begin{abstract}
The sample compression conjecture is a longstanding open problem in learning theory, originating in the work of Littlestone and Warmuth in the 1980s. At its heart is a connection between simplification and generalization: if a predictor consistent with a training sample is defined using only a small part of that sample, then it is likely to generalize. Sample compression schemes formalize this idea, and many natural learning algorithms can be analyzed this way.

The sample compression conjecture of Floyd and Warmuth (1995) asserts that
every binary concept class of VC dimension $d$ admits a sample compression
scheme of size $O(d)$; a weaker conjecture asks only for size polynomial
in $d$. Both have remained open: the first bound depending only on $d$, due
to Moran and Yehudayoff (2016), is exponential in $d$. We prove the weaker
conjecture by constructing a sample compression scheme of size
$O(d^2 \log^2 d)$.
\end{abstract}

\section{Introduction}\label{sec:intro}

Learning has two complementary aspects: generalization and simplification. A scientist who observes the outcomes of many experiments may seek to predict the outcome of a new one; this is generalization. The same scientist may also seek to explain the observed outcomes in terms of a small number of informative experiments, from which the remaining observations can be recovered by reasoning; this is simplification. These perspectives are connected by a classical guiding principle: Occam's razor suggests favoring simpler explanations among those consistent with the observations. In learning, this principle motivates the intuition that a concise explanation of observed data may also generalize well to new examples. Understanding when and how this intuition can be made precise is a fundamental question about learning. Simplification is also of particular interest in modern machine learning, where models that generalize well are often difficult to interpret.

In learning theory, generalization is captured by Valiant's PAC learning framework~\cite{Val84}, with VC theory providing a fundamental characterization of learnability~\cite{VC71,BEHW89}. Sample compression offers a complementary framework for studying simplification.

Littlestone and Warmuth introduced sample compression in 1986~\cite{LW86} as a way to derive generalization bounds that reflect the behavior of natural learning algorithms. In many geometric problems, the concept returned by a learning algorithm is determined by only a few examples. For instance, the smallest axis-aligned rectangle containing all positive examples in the plane is determined by at most four of them: one for each of its sides. Once these examples are known, the rectangle can be reconstructed without remembering the rest of the sample. This observation suggests a way to understand generalization through the amount of data needed to specify the learner's output. Indeed, and as discussed in more detail below, Littlestone and Warmuth showed that such algorithms generalize in Valiant's PAC setting.
\begin{figure}[htbp]
\centering
\begin{tikzpicture}[x=0.85cm,y=0.85cm,>=Latex,font=\small,sample/.style={circle,fill=black,inner sep=1.6pt},retained/.style={circle,draw=blue!65!black,line width=1pt,fill=white,minimum size=7pt,inner sep=0pt}]
\draw[->,gray] (0,0)--(8.2,0) node[right] {$x_1$};
\draw[->,gray] (0,0)--(0,5.3) node[above] {$x_2$};
\fill[blue!4] (1,1) rectangle (7,4.5);
\draw[blue!65!black,thick] (1,1) rectangle (7,4.5);
\foreach \p in {(1.6,1.7),(2.1,3.5),(2.7,2.4),(3.1,4.0),(3.8,1.8),(4.2,3.2),(4.9,2.4),(5.5,3.8),(5.8,1.6),(6.4,2.5)} \node[sample] at \p {};
\foreach \x/\y in {0.55/3.8,1.8/4.95,6.2/0.45,7.65/1.5,7.7/4.7}{
\draw[gray,thick] (\x-0.07,\y-0.07)--(\x+0.07,\y+0.07);
\draw[gray,thick] (\x-0.07,\y+0.07)--(\x+0.07,\y-0.07);}
\foreach \p/\name/\where in {(1,2.5)/z_L/left,(7,3)/z_R/right,(3.2,1)/z_B/below,(4.5,4.5)/z_T/above}{
\node[retained] at \p {};
\node[sample] at \p {};
\node[\where=5pt,blue!65!black] at \p {$\name$};}
\node[font=\footnotesize] at (4,6.0) {Solid dots: positive examples.\quad Crosses: negative examples.};
\end{tikzpicture}
\caption{Compression for axis-aligned rectangles. The compressor retains the circled positive examples $T$, at most one for each side. The reconstructor outputs the smallest axis-aligned rectangle containing $T$. For a sample realizable by an axis-aligned rectangle, this rectangle contains every positive example and excludes every negative example.}
\label{fig:rectangle-compression}
\end{figure}
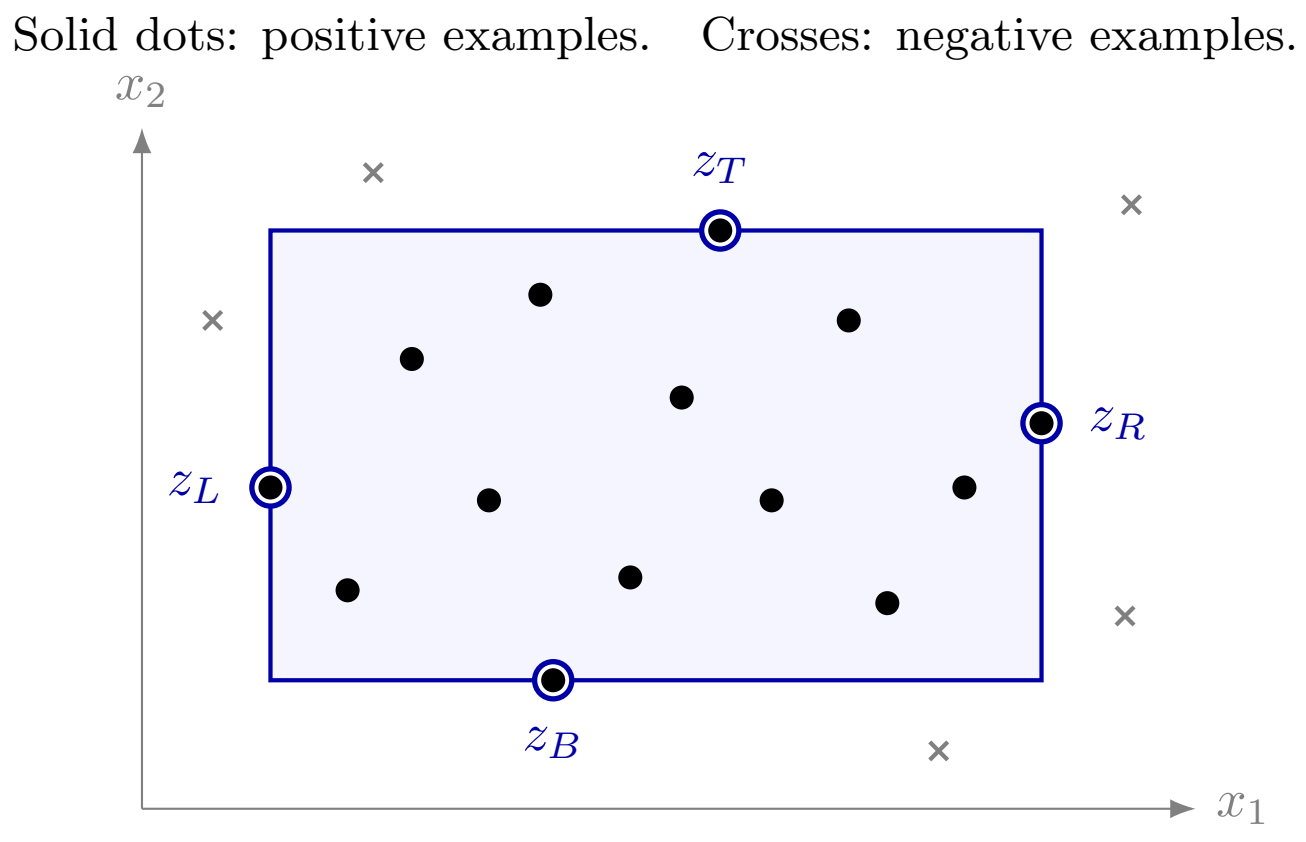
A sample compression scheme formalizes the idea that a few examples can determine a prediction rule consistent with the entire sample. Such a scheme can be viewed as a protocol between two players: a compressor and a reconstructor.
The compressor receives a labeled dataset and sends the reconstructor a small subset of its examples, possibly accompanied by a short string of side-information bits. From this message alone, the reconstructor must produce a concept consistent with every example in the original dataset, including those it never received.

More formally, let $\cX$ be a domain of examples and let $\cC\subseteq\{0,1\}^{\cX}$ be a concept class. A finite labeled sample $S=((x_1,y_1),\ldots,(x_m,y_m))$ is $\cC$-realizable if some $c\in\cC$ satisfies $c(x_i)=y_i$ for every $i$. A \emph{sample compression scheme of size $(k,b)$} for $\cC$ consists of a compressor $\kappa$ and a reconstructor~$\rho$. Given any such sample, $\kappa(S)=(T,\sigma)$ selects an unordered subset $T\subseteq\{(x_1,y_1),\ldots,(x_m,y_m)\}$ with $|T|\leq k$ and a bit string $\sigma\in\{0,1\}^{\leq b}$. Given only $T$ and $\sigma$, the reconstructor outputs a concept $h=\rho(T,\sigma):\cX\to\{0,1\}$ satisfying $h(x_i)=y_i$ for every $i$, including examples outside $T$. The two bounds $k$ and $b$ hold uniformly over all realizable samples, independently of their size.

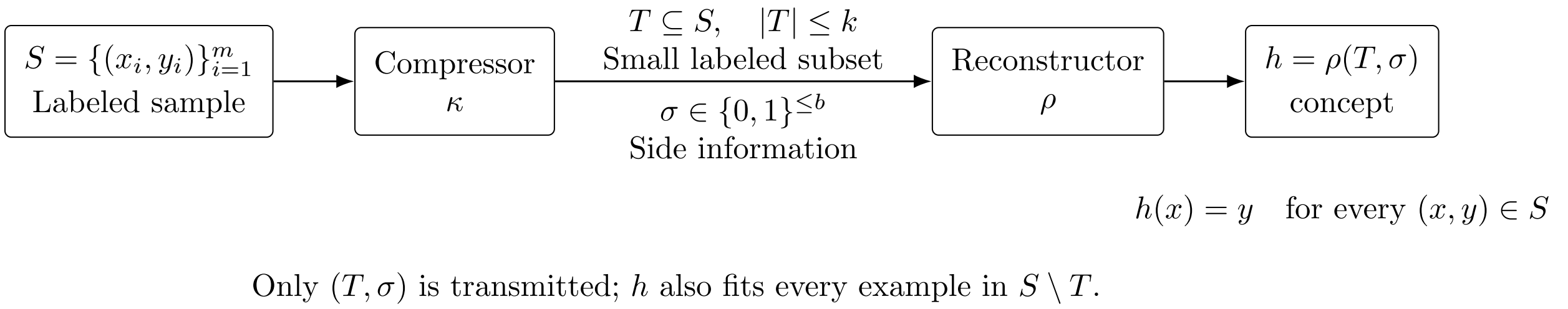
\begin{figure}[htbp]
\centering
\resizebox{\linewidth}{!}{%
\begin{tikzpicture}[
  >=Latex, font=\small, node distance=9mm,
  block/.style={draw, rounded corners=2pt, align=center, minimum height=12mm, inner sep=6pt},
  flow/.style={->, semithick}
]
  \node[block] (sample) {$S=\{(x_i,y_i)\}_{i=1}^{m}$\\[2pt]Labeled sample};
  \node[block, right=of sample] (compressor) {Compressor\\$\kappa$};
  \node[block, right=42mm of compressor] (reconstructor) {Reconstructor\\$\rho$};
  \node[block, right=of reconstructor] (concept) {$h=\rho(T,\sigma)$\\[2pt]concept};
  \draw[flow] (sample) -- (compressor);
  \draw[flow] (compressor) -- node[above, align=center] {$T\subseteq S,\quad |T|\leq k$\\Small labeled subset} node[below, align=center] {$\sigma\in\{0,1\}^{\leq b}$\\Side information} (reconstructor);
  \draw[flow] (reconstructor) -- (concept);
  \node[below=5mm of concept, align=center] (consistency) {$h(x)=y\quad\text{for every }(x,y)\in S$};
  \node[below=17mm of compressor, anchor=west, xshift=-24mm] {Only $(T,\sigma)$ is transmitted; $h$ also fits every example in $S\setminus T$.};
\end{tikzpicture}%
}
\caption{A sample compression scheme. The reconstructed concept is consistent with the entire sample, not only the retained examples.}
\label{fig:compression-protocol}
\end{figure}

\paragraph{Compression implies generalization.}
Littlestone and Warmuth~\cite{LW86} proved that bounded-size compression guarantees PAC learnability. For any fixed choice of compression indices and side information, the reconstructed concept is independent of the remaining examples. A union bound over these choices yields the generalization guarantee. Quantitatively, a compression scheme of size $(k,b)$ yields, for target generalization error $\varepsilon$, PAC sample complexity $\wt O((k+b)/\varepsilon)$~\cite{LW86,FW95,GHS05}; so-called \emph{stable sample compression schemes} admit sharper bounds of order $O(k/\varepsilon)$~\cite{Zhi17,BHMZ20,HK21}, with confidence dependence suppressed here. 

This is an algorithm-dependent expression of Occam's razor~\cite{SB14}. 
This viewpoint extends classical minimum-description-length and bit-length Occam bounds~\cite{Ris78,BEHW87}, which use a fixed binary encoding of concepts. In the compression framework, such descriptions correspond to retaining no examples and transmitting the entire description as side information. Allowing the description to be anchored in the sample itself is substantially more expressive: bounded-length binary descriptions specify only finitely many concepts, whereas a bounded number of examples can encode concepts from an infinite class, such as thresholds on $\mathbb R$. Geometric compression and margin-based learners illustrate this advantage; for example, Perceptron-style learners under margin $\gamma$ on unit-norm instances retain $O(1/\gamma^2)$ examples even in an infinite-dimensional Hilbert space~\cite{GHS05,HK21}.

The reach of compression-based generalization goes beyond realizable binary classification. It extends to agnostic learning~\cite{DMY16} and real-valued regression~\cite{HKS19,AHKS24}, and---allowing compression size to grow polylogarithmically with the sample size where needed---to settings where uniform convergence can fail, including partial concept classes~\cite{AHHM21}, multiclass classification over infinite label spaces~\cite{DMY16,BCDMY22}, adversarially robust classification~\cite{MHS19}, and density estimation for mixtures of Gaussians~\cite{ABDHMP20}.

\paragraph{From learnability to compression.}
Having established that compressibility implies PAC learnability, Littlestone and Warmuth~\cite{LW86} posed the converse question for binary classification: does every PAC-learnable class, equivalently every concept class of finite VC dimension $d$, admit a sample compression scheme whose size depends only on $d$? Floyd and Warmuth~\cite{FW95} further conjectured that $O(d)$ examples suffice, a question now known as the \emph{sample compression conjecture}. 

Conceptually, this asks how far Occam's principle can be reversed: if a class can be learned from data, can every realizable sample also be simplified into a bounded number of its examples, from which all its labels can be recovered?

Allowing dependence on the sample size $m$, {Freund~\cite{Fre95} and Helmbold and Warmuth~\cite{HW95}} used boosting to obtain schemes retaining $O(d\log m)$ examples (plus side information, as noted by Floyd and Warmuth~\cite[Section~9.1]{FW95}), while schemes of exact size $d$ were constructed for structured families such as maximum and extremal classes~\cite{FW95,MW16}. For general VC classes, the existence of any sample-independent bound $f(d)$ remained open for thirty years, until Moran and Yehudayoff~\cite{MY16} constructed a scheme of size $2^{O(d)}$.

This exponential bound leaves a substantial quantitative gap between learnability and compression: the optimal PAC sample complexity of VC-$d$ classes is of order $d/\varepsilon$ for constant confidence~\cite{BEHW89,Han16}, whereas the generalization guarantee obtained from the exponential compression scheme is exponentially larger in $d$. Even the weaker question of whether every VC class admits a polynomial-size compression scheme has remained a central open problem~\cite{FW95,War03,MY16}.

In this work, we answer the polynomial sample compression question in the affirmative, obtaining the first asymptotic improvement over~\cite{MY16} for general VC classes and achieving near-quadratic compression size $\wt O(d^2)$. Throughout, $\log x$ in asymptotic bounds denotes $\log_2\max\{x,2\}$.

\begin{samepage}
\begin{theorem}[Main Theorem]\label{thm:main}
Every concept class $\cC \subseteq \{0,1\}^{\cX}$ of VC dimension $d \ge 1$ admits a labeled sample compression scheme of size $k = O(d^2 \log^2 d)$ examples and $b = O(d \log^2 d)$ side-information bits.
\end{theorem}
\end{samepage}

Our bounds concern the size of the compressed representation, not the computational complexity of finding or reconstructing it.

\begin{remark}\label{rem:variants}
Our result and proof approach have several properties and implications; we list some of them below:
\begin{enumerate}[label=(\alph*), leftmargin=1.8em, itemsep=1pt, topsep=2pt]
  \item \textbf{Weighted majority scheme:} The reconstructed concept $h = \rho(T, \sigma)$ is always a weighted majority vote of at most $(k + 1)^d$ concepts from $\cC$ (placing $> 1/2$ weight on the true label at every sample point; see Remark~\ref{rem:infinite} for infinite domains).
  \item \textbf{Unlabeled vs.\ labeled examples:} Our scheme requires only $O(d \log d)$ labeled examples (and $O(d \log^2 d)$ bits of side information), while the remaining $O(d^2 \log^2 d)$ examples are \emph{unlabeled} points (likewise preserved on infinite domains via Remark~\ref{rem:infinite}).
  \item \textbf{Agnostic sample compression.}
Our result also implies an agnostic sample compression scheme of the same size. Indeed, and as observed by David, Moran, and Yehudayoff~\cite{DMY16}, such a scheme can be obtained by applying a realizable sample compression scheme to the subsample on which an empirical risk minimizer $c^* \in \mathcal{C}$ agrees with the labels. The resulting predictor is consistent with this subsample and therefore makes no more errors on the original sample than~$c^*$. 
\end{enumerate}
\end{remark}

\subsection{Proof Overview}\label{sec:overview}

Let us recall what we need to prove. Given a concept class $\cC$ of VC dimension $d$, we seek a sample compression scheme with the bounds stated in the theorem. Given a realizable labeled sample $S$, the compressor stores a small number of examples and a small amount of side information. From this encoding, the reconstructor must recover a concept consistent with the entire sample. Without loss of generality, we assume that $\cC$ is finite, by the compactness theorem for sample compression of Ben-David and Litman~\cite{BDL98}.

As in Moran and Yehudayoff's scheme~\cite{MY16}, our approach is to reconstruct a weighted majority vote of concepts from $\cC$. We can view the normalized weights as a distribution over concepts. If, at each sample point, a concept drawn from this distribution predicts the correct label with probability greater than $1/2$, the weighted vote agrees with every label in the sample.

Moran and Yehudayoff use the minimax theorem to obtain a distribution over concepts with low error at each sample point. They then sparsify this distribution by sampling a short list of concepts whose majority vote remains consistent with the sample. A similar sampling argument is used in~\cite{LY94,LMM03} to sparsify mixed strategies. In that setting, a union bound gives a logarithmic dependence on the number of the opponent's pure strategies. Moran and Yehudayoff use uniform convergence for the dual class to replace this logarithmic dependence with a bound in terms of the dual VC dimension. \emph{We avoid the dependence on the dual VC dimension, which can be exponential in $d$, by encoding the set of concepts used in the weighted majority vote with a small set of unlabeled examples.}

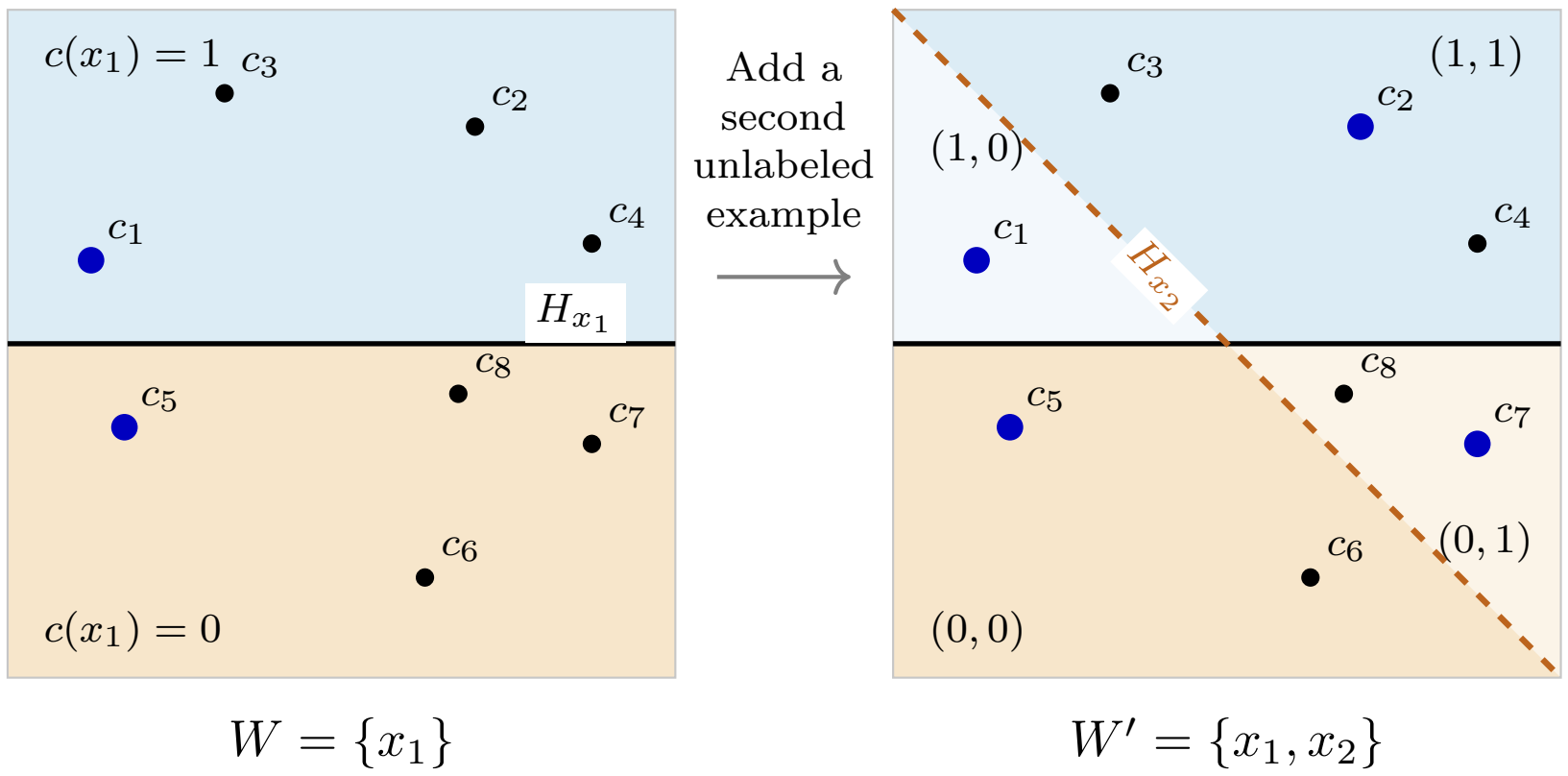
\begin{figure}[!t] 
\centering
\begingroup
\definecolor{poolink}{HTML}{000000}
\definecolor{poolblue}{HTML}{DCECF5}
\definecolor{poolorange}{HTML}{F7E6CB}
\definecolor{poolcut}{HTML}{BC641D}
\def\poolpoints{\foreach \i/\a/\b in {1/-1.5/.5,2/.8/1.3,3/-.7/1.5,4/1.5/.6,5/-1.3/-.5,6/.5/-1.4,7/1.5/-.6,8/.7/-.3}{\fill[poolink] (\a,\b) circle (1.8pt);\node[font=\scriptsize,poolink,anchor=south west,inner sep=3pt] at (\a,\b) {$c_{\i}$};}}
\def\poolrep#1#2{\fill[blue!75!black] (#1,#2) circle (2.6pt);}
\begin{tikzpicture}[x=1.15cm,y=1.15cm,font=\small]
\begin{scope}
\fill[poolblue] (-2,0) rectangle (2,2);
\fill[poolorange] (-2,-2) rectangle (2,0);
\draw[gray!45] (-2,-2) rectangle (2,2);
\draw[poolink,line width=1pt] (-2,0)--(2,0);
\node[above,font=\scriptsize,fill=white,inner sep=2pt] at (1.4,0) {$H_{x_1}$};
\poolpoints \poolrep{-1.5}{.5} \poolrep{-1.3}{-.5}
\node[font=\scriptsize,anchor=north west] at (-1.9,1.96) {$c(x_1)=1$};
\node[font=\scriptsize,anchor=south west] at (-1.9,-1.96) {$c(x_1)=0$};
\node[below] at (0,-2.12) {$W=\{x_1\}$};

\end{scope}
\draw[->,thick,gray] (2.25,0.4)--(3.05,0.4);
\node[align=center,font=\scriptsize] at (2.65,1.2) {Add a\\second\\unlabeled\\example};
\begin{scope}[xshift=6.1cm]
\fill[poolblue] (-2,0)--(-2,2)--(2,2)--(2,0)--cycle;
\fill[poolblue!35!white] (-2,0)--(-2,2)--(0,0)--cycle;
\fill[poolorange] (-2,-2)--(-2,0)--(2,0)--(2,-2)--cycle;
\fill[poolorange!45!white] (0,0)--(2,0)--(2,-2)--cycle;
\draw[gray!45] (-2,-2) rectangle (2,2);
\draw[poolink,line width=1pt] (-2,0)--(2,0);
\draw[poolcut,line width=1.2pt,dashed] (-2,2)--(2,-2);
\poolpoints \poolrep{-1.5}{.5} \poolrep{.8}{1.3} \poolrep{-1.3}{-.5} \poolrep{1.5}{-.6}
\node[font=\scriptsize,anchor=north west] at (-1.9,1.4) {$(1,0)$};
\node[font=\scriptsize,anchor=north east] at (1.9,1.96) {$(1,1)$};
\node[font=\scriptsize,anchor=south west] at (-1.9,-1.96) {$(0,0)$};
\node[font=\scriptsize,anchor=north east] at (1.95,-.95) {$(0,1)$};
\node[poolcut,font=\scriptsize,anchor=west,rotate=-45,fill=white,inner sep=1pt] at (-.6,.6) {$H_{x_2}$};
\node[below] at (0,-2.12) {$W'=\{x_1,x_2\}$};

\end{scope}
\end{tikzpicture}
\endgroup
\caption{The black dots represent linear classifiers as points in the dual space. The blue dots mark the representatives of the occupied cells. On the left, the line corresponding to one unlabeled example separates classifiers that label it 0 from those that label it 1. On the right, a second unlabeled example adds a line that refines the partition.}
\label{fig:partition-refinement}
\end{figure}  \paragraph{Encoding and Reconstruction.} Recall that our encoding contains $O(d^2\log^2 d)$ unlabeled examples and $O(d\log d)$ labeled examples. The unlabeled examples specify which concepts participate in the vote, and the labeled examples specify their weights.

How do the unlabeled examples specify the concepts? Fix a linear order on $\cC$. For every realizable labeling of the unlabeled examples, take the first concept in that order that realizes it. The crucial property is that the resulting set of representatives admits a distribution $q$ with low error at every sample point:
\begin{equation}\label{eq:overview-pointwise}
\Pr_{h\sim q}[h(x)\ne y]\le \tfrac14
\qquad\text{for every }(x,y)\in S \,.
\end{equation}

We note that the constant $1/4$ is chosen for concreteness; in fact, any constant bounded away from $1/2$ would work. 
For now, assume that we have a set of unlabeled examples whose representatives admit a distribution satisfying~\eqref{eq:overview-pointwise}. We first explain how to use these representatives to reconstruct a weighted majority vote consistent with the sample. In the next part, we show how to choose the unlabeled examples so that their representatives have this property.

The pointwise error bound in~\eqref{eq:overview-pointwise} is what makes the weighted majority step work: the compressor runs 
the multiplicative weights update (MWU) algorithm
on the representatives. In each round it selects a point from the full sample on which the current weights assign the most mass to the wrong label. The standard analysis shows that after $O(d\log d)$ rounds the vote formed from the average of the round-by-round weights agrees with every sample label. The compressor stores the labeled examples used in these updates (these are the additional $O(d\log d)$ labeled examples used in our scheme). The reconstructor runs the algorithm on those examples and returns this weighted vote.

  \paragraph{Choosing the unlabeled examples.} The most technical part of the proof is choosing the unlabeled examples so that their representatives admit a distribution satisfying~\eqref{eq:overview-pointwise}. To visualize the construction, think of the concepts as linear classifiers (this is only for illustration; the argument applies to any finite concept class). Geometrically, each linear classifier can be represented as a point in the dual parameter space, and each unlabeled example defines a hyperplane, shown as a line in Figure~\ref{fig:partition-refinement}. This hyperplane separates the classifiers that label the example $0$ from those that label it $1$.

We start with a single cell and add a few hyperplanes in each round to refine the partition. The classifiers in each cell agree on all the examples chosen so far, and the first classifier in the fixed order is the cell's representative. We continue until the set of representatives admits a distribution satisfying~\eqref{eq:overview-pointwise}.

We maintain a pair $(\mathcal P_t,\mu_t)$: the hyperplanes chosen so far determine the partition $\mathcal P_t$, and $\mu_t$ is a distribution over the entire class $\cC$. Initially, there are no hyperplanes, so $\mathcal P_0$ has a single cell, and $\mu_0$ puts all its mass on the target concept $c^*$. If $c^*$ is first in the fixed order, it is the representative of this cell, and $\mu_0$ already satisfies~\eqref{eq:overview-pointwise}, so we stop.
Otherwise, we update the distribution and refine the partition by adding hyperplanes. We add $O(d\log d)$ unlabeled examples per continuing round and show that $O(d\log d)$ such rounds suffice; the final set contains $O(d^2\log^2 d)$ unlabeled examples.
An important invariant of the process is that the entropy of the distribution induced by the {cell masses} increases {in each continuing round by at least a universal constant} (these masses indeed form a {distribution} since the cells partition the space). Note that the initial entropy of the cell masses is $0$, since the only cell has mass $1$ and that the entropy is always upper bounded by the logarithm of the number of {occupied} cells, and that the latter is upper bounded by the Sauer--Shelah--Perles lemma.

As the construction proceeds, we keep the pointwise error at stage $t$ at most $t\delta$. 
We stop when the set $A$ of current representatives has mass $\mu(A)\ge 1/4$. Our choice of $\delta$ ensures that the pointwise error under $\mu$ is at most $1/16$ at termination. Thus $q=\mu(\cdot\mid A)$ satisfies~\eqref{eq:overview-pointwise}, since for every $(x,y)\in S$,
\[
\Pr_{h\sim q}[h(x)\ne y]
=\frac{\Pr_{h\sim\mu}[h(x)\ne y,\ h\in A]}{\mu(A)}
\le \frac{1/16}{1/4}=\frac14.
\]

Let $(\mathcal P_t,\mu_t)$ be the current pair of a partition and distribution.
To construct the next pair $(\mathcal P_{t+1},\mu_{t+1})$, we first use a minimax argument to choose $\mu_{t+1}$ and then use $\mu_{t+1}$ to derive $\mathcal{P}_{t+1}$.
The derivation of $(\mathcal P_{t+1},\mu_{t+1})$ is the most involved step in the proof and we discuss it in further detail at the end of this overview. For now, we just describe the necessary properties of the construction and how they imply an upper bound on the number of unlabeled examples.
The distribution $\mu_{t+1}$ satisfies \[ \mu_{t+1}(C)=\mu_t(C) \qquad\text{for every }C\in\mathcal P_t, \] and \[ \Pr_{h\sim\mu_{t+1}}[h(x)\ne y]\le(t+1)\delta \qquad\text{for every }(x,y)\in S. \] Thus, changing the distribution preserves the mass of every current cell and therefore preserves the entropy of the current partition.

We now hold $\mu_{t+1}$ fixed and construct the refined partition $\mathcal P_{t+1}$; there are two possibilities: either a final batch of $O(d^2\log^2 d)$ unlabeled examples produces a partition whose representatives carry mass at least $1/4$, in which case conditioning on them gives a distribution satisfying~\eqref{eq:overview-pointwise} and we stop; or a batch of $O(d\log d)$ unlabeled examples increases the entropy of the cell masses by at least a universal constant $\gamma>0$. Geometrically, these examples add hyperplanes that split the current cells. In the continuing case, the refined partition and the fixed distribution form the next pair $(\mathcal P_{t+1},\mu_{t+1})$. 

To bound the number of continuing rounds, let $R=\Theta(d\log d)$ and set $\delta=1/(16R)$. Let $B=O(1/\delta)=O(d\log d)$ bound the number of examples added per continuing round, and let $H_r$ be the entropy after $r\ge1$ such rounds. The entropy increase and the Sauer--Shelah--Perles bound on the number of occupied cells give \[ \gamma r\le H_r\le d\log_2(eBr). \] For a sufficiently large constant in $R$, these inequalities force termination within $R$ rounds. The continuing rounds use $O(d^2\log^2 d)$ unlabeled examples in total. The possible final batch uses $O((d/\delta)\log(1/\delta))=O(d^2\log^2 d)$ additional examples, so the same bound holds for the entire construction.

\paragraph{Choosing the next distribution and partition.} We now explain the ideas behind the two steps above. Let $\mathcal K_t$ be the polytope of distributions over $\cC$ that preserve the current cell masses and have pointwise error at most $(t+1)\delta$. 
Formally, \[ \mathcal K_t=\left\{\nu\in\Delta(\cC): \begin{array}{ll} \nu(C)=\mu_t(C) & \text{for every }C\in\mathcal P_t,\\[2pt] \Pr_{h\sim\nu}[h(x)\ne y]\le(t+1)\delta & \text{for every }(x,y)\in S \end{array} \right\}, \] where $\Delta(\cC)$ denotes the set of probability distributions over $\cC$.

We define a zero-sum game in order to choose $\mu_{t+1}$ within this polytope. 
In this game, both players' strategies are the distributions in $\mathcal K_t$. The payoff is determined as follows. First, a common cell $C$ is chosen with probability $\mu_t(C)$. Each player then independently draws a concept from its chosen distribution conditioned on $C$. The earlier concept in the fixed order wins, with ties broken uniformly. By symmetry and the minimax theorem, the game has value $1/2$ and admits a symmetric equilibrium. We choose $\mu_{t+1}$ to be an optimal distribution in this equilibrium. Thus, no feasible distribution in $\mathcal K_t$ beats it with probability greater than $1/2$.

We now fix the opponent's strategy to $\mu_{t+1}$ and consider the
problem of finding a best response within $\mathcal K_t$. This is a
linear program, and the equilibrium property guarantees that
$\mu_{t+1}$ itself is an optimal solution. If the representatives already carry at least $1/4$ of the mass of $\mu_{t+1}$, we are done. Otherwise, linear programming duality,
together with the $\delta$-slack of $\mu_t$, yields a distribution
$p^*$ over sample points. The key property of $p^*$ is that the hyperplane corresponding to a random point drawn from it separates
concepts in the same cell with probability bounded below in terms of
their distance in the fixed linear order. The distance here is measured by the
conditional $\mu_{t+1}$-mass between them. This property allows us to choose a small set of hyperplanes that refines the partition in the desired way.

This yields the two refinement cases described above. If at least half the cell mass lies in cells with a concept carrying more than half of that cell's mass, a $(\delta/2)$-net under $p^*$ separates these concepts from all their earlier predecessors. They become representatives of the refined cells and together carry mass greater than $1/4$. Otherwise, $O(1/\delta)$ examples drawn from $p^*$ split the spread-out cells enough to increase their total cell-mass entropy by a universal constant. Deriving $p^*$ from linear programming duality and analyzing these two cases are the main technical steps; the details appear in Section~\ref{sec:dichotomy}.

\subsection{Related Work}\label{sec:related}

While our overall compression scheme is new, its individual components connect naturally to classical ideas across learning theory, algorithmic game theory, optimization, and combinatorics. Below, we survey some of these connections and related works.

\paragraph{Upper bounds for structured classes and general VC classes.}
Following Littlestone and Warmuth~\cite{LW86}, linear-size compression schemes were constructed for many structured families. Floyd and Warmuth~\cite{FW95} gave labeled schemes of size $d$ for \emph{maximum} classes. Maximum classes also admit unlabeled compression schemes of size $d$~\cite{CCMW22}, and Moran and Warmuth~\cite{MW16} established size-$d$ labeled schemes for all \emph{extremal} classes. Size-$O(d)$ schemes are likewise known for intersection-closed classes~\cite{HSW90,FW95} and halfspaces in $\mathbb{R}^d$ via stable compression~\cite{FW95,Zhi17,BHMZ20,HK21}, while sample-independent schemes parameterized by auxiliary geometric or model-theoretic invariants were established for classes of bounded variability~\cite{BDL98}, Helly-type classes~\cite{BHMZ20}, and definable families in stable, o-minimal, and NIP structures~\cite{JL10,LS13,CS13,CS15}. For arbitrary VC classes, early bounds depended on the sample size $m$ or the class cardinality $|\cC|$, retaining $O(d \log m)$ examples (plus side information) via boosting~\cite{Fre95,HW95} and $O(d 2^d \log\log |\cC|)$~\cite{MSWY15}. The first bound depending solely on $d$ was given by Moran and Yehudayoff~\cite{MY16}, who devised sample compression schemes with exponential dependence on $d$, subsequently extended and reduced to agnostic, multiclass, real-valued, and adversarially robust settings~\cite{DMY16,MHS19,HKS19,AHKS24,AHR25}.

\paragraph{Lower bounds and structural obstructions.}
On the lower bound side, simple counting shows that worst-case classes of VC dimension $d$ (such as subsets of $\mathbb{N}$ of size at most $d$) require $k \ge d$ examples regardless of side-information bits, while every class of VC dimension $d$ requires $k \ge d/5$ without side bits (or $k + b = \Omega(d)$ with side bits) even on a single shattered $d$-tuple~\cite{FW95}. While no binary class is known to require strictly more than $d$ \emph{labeled} examples, P{\'a}lv{\"o}lgyi and Tardos~\cite{PT20} disproved Kuzmin and Warmuth's analogous size-$d$ conjecture for \emph{unlabeled} compression, exhibiting VC classes whose smallest unlabeled scheme strictly exceeds $d$ (e.g., size $3$ at $d = 2$). On the methodological side, since maximum and extremal classes of VC dimension $k$ admit size-$k$ compression schemes~\cite{FW95,MW16}, Floyd and Warmuth~\cite{FW95} suggested compressing a general VC-$d$ class $\cC$ by embedding it into a maximum (or extremal) class $\cC'$ of small VC dimension $k$ and applying the compression scheme of $\cC'$. Chase, Chornomaz, Hanneke, Moran, and Yehudayoff~\cite{CCHMY24} proved an exponential barrier to this approach: embedding $\cC$ into $\cC'$ preserves dual shattering ($d^*(\cC) \le d^*(\cC')$), whereas every extremal class $\cC'$ of VC dimension $k$ has dual VC dimension $d^*(\cC') \le 2k + 1$---forcing $k = \Omega(d^*(\cC)) = 2^{\Omega(d)}$ whenever $\cC$ has exponential dual VC dimension.

\paragraph{Compression beyond binary labels.}
Beyond total binary concepts, the relationship between learnability, uniform convergence, and sample compression changes substantially. For multiclass concept classes $\cH \subseteq \mathcal{Y}^{\cX}$, uniform convergence is characterized by finiteness of the \emph{graph dimension} $d_{\mathrm{G}}$~\cite{BDCHL95,DSBS15}; in particular, over a finite label space, every PAC-learnable class satisfies $d_{\mathrm{G}} = O(d_{\mathrm{N}} \log |\mathcal{Y}|)$, where $d_{\mathrm{N}}$ denotes the Natarajan dimension~\cite{Nat89,BDCHL95}. Whenever $d_{\mathrm{G}}$ is finite (even for infinite $\mathcal{Y}$), reducing to the binary graph class yields sample-independent compression of size $2^{O(d_{\mathrm{G}})}$~\cite{DMY16}; because our binary scheme outputs a weighted majority of concepts from $\cC$ (Remark~\ref{rem:variants}(a)), plugging Theorem~\ref{thm:main} into the reductions of~\cite{DMY16,AHR25} improves this to $O(d_{\mathrm{G}}^2 \log^2 d_{\mathrm{G}})$ examples and $O(d_{\mathrm{G}} \log^2 d_{\mathrm{G}})$ side-information bits (and similarly gives $\wt{O}(d_{\mathrm{P}}^2)$-size $\varepsilon$-approximate compression, independent of $\varepsilon$, for $[0,1]$-valued classes of pseudo-dimension $d_{\mathrm{P}}$~\cite{AHR25}). Over infinite label spaces, however, uniform convergence is no longer necessary for learnability, and neither the graph dimension nor the Natarajan dimension characterizes PAC learnability~\cite{DS14,DSBS15,BCDMY22}. Instead, Brukhim, Carmon, Dinur, Moran, and Yehudayoff~\cite{BCDMY22} proved that multiclass PAC learnability is characterized by the Daniely--Shalev-Shwartz (DS) dimension $d_{\mathrm{DS}}$~\cite{DS14}, establishing learnability via compression schemes whose size grows polylogarithmically with the sample size $m$. In sharp contrast to the binary case~\cite{MY16}, Pabbaraju~\cite{Pab24} proved that this dependence on $m$ cannot be eliminated: there exists a class over $\mathcal{Y} = \mathbb{N}$ with $d_{\mathrm{DS}} = 1$ (hence PAC learnable) that admits no bounded-size compression scheme, as any scheme on samples of size $m$ requires size $\Omega((\log m)^{1 - o(1)})$. Thus, multiclass PAC learnability does \emph{not} imply sample-independent compression. Analogous separations arise for \emph{partial} binary concept classes, where classes of VC dimension $1$ are PAC learnable yet can fail to admit any bounded-size compression scheme~\cite{AHHM21}, and for \emph{list} PAC learning, where Hanneke, Moran, and Waknine~\cite{HMW24} showed that even over three labels (where uniform convergence holds), $2$-list learnable classes need not admit any bounded-size $2$-list compression scheme.

\paragraph{Boosting as compression and Multiplicative Weights.}
Using ensemble updates to build sample-dependent compression schemes (retaining $O(d \log m)$ examples, plus side information) goes back to Freund~\cite{Fre95} and Helmbold and Warmuth~\cite{HW95}, building on boosting~\cite{Sch90}, while Moran and Yehudayoff~\cite{MY16} combined von Neumann's minimax theorem~\cite{vN28} with dual-VC uniform convergence to eliminate $\log m$ at the price of $d^*$. Lemma~\ref{lem:mwu} instead applies the regret bound of Multiplicative Weights~\cite{LW94,FS97,CBL06} to the implicitly transmitted $\Pool(W)$, turning its $(|W| + 1)^d$ cardinality into a logarithmic $O(d \log(|W| + 2))$ overhead of labeled examples.

\paragraph{Canonical cell representatives and recursive teaching.}
Using a fixed sample-independent order to break ties deterministically is standard in sample compression~\cite{FW95,MY16} (coupled with Ben-David and Litman's compactness reduction to finite domains~\cite{BDL98}). Decoding a concept as $\min_\prec C_\tau$ from a small defining set $W$ also echoes collusion-free teaching models such as the \emph{recursive} and \emph{no-clash teaching dimensions} ($\mathrm{RTD}$, $\mathrm{NCTD}$)~\cite{DFSZ14,KSZ19}, with Hu, Wu, Li, and Wang~\cite{HWLW17} proving $\mathrm{RTD}(\cC) = O(d^2)$. Crucially, however, $\mathrm{RTD}$ tailors $\prec$ carefully to $\cC$ and selects teaching points from the \emph{entire domain} $\cX$, which need not appear in a sample $X \subseteq \cX$. Here $\prec$ is completely arbitrary, and $\Pool(W)$ is defined via $\min_\prec(C_\tau)$ across all realizable labelings $\tau$ of $W \subseteq X$ by $\cC$.

\paragraph{Maximal lotteries.}
Pushing a distribution downward along an ordinal ranking $\prec$ via the constant-sum head-to-head payoff $M(\nu, \eta) = \Pr[\eta \prec \nu] + \frac{1}{2}\Pr[\eta = \nu]$ originates in probabilistic social choice as \emph{maximal lotteries} (Kreweras~\cite{Kre65}; Fishburn~\cite{Fis84}) and appears in learning theory as the \emph{von Neumann winner} of dueling bandits~\cite{DHSSZ15}. In Sections~\ref{subsec:qem} and~\ref{subsec:qem-proof} we define an appropriate polytope and run this game over it to obtain a symmetric saddle point $(\mu^*, \mu^*)$ and, via its first-order optimality condition, the distribution $p^*$ described above.

\paragraph{Slater's condition and $\varepsilon$-nets.}
An important technical step towards deriving the distribution $p^*$ is bounding the $\ell_1$-norm $\Lambda = \sum_x \lambda_x \le \frac{1}{2\delta}$ of the dual multipliers using a feasible point $\mu_r$ with uniform slack $\delta$, via the classical Slater bound in convex optimization~\cite{Uza58}. Increasing the error budget from $r\delta$ to $(r + 1)\delta$ ensures that each round's iterate $\mu_r$ satisfies Slater's condition for the next round. Moreover, complementary slackness in Section~\ref{subsec:qem-proof} is asymmetric---requiring $\mu^*(m_\tau) > 0$ only on a heavy concept $m_\tau$ and not on earlier concepts $b \prec m_\tau$ (which may carry $\mu^*(b) = 0$)---and because separating $b \prec m_\tau$ from $m_\tau$ only requires hitting a single disagreement point rather than approximating disagreement frequencies, we can terminate using a $(\delta/2)$-net~\cite{HW87} of size $\wt{O}(d/\delta)$ for $\cC \oplus \cC$ rather than a $(\delta/2)$-approximation of size $\wt{O}(d/\delta^2)$.

\paragraph{Cell-entropy potential.}
Partitioning a bounded-VC family by restrictions $c|_W$ to a small witness set $W$ so that the Sauer--Shelah--Perles lemma yields at most $(|W|+1)^d$ cells rather than $2^{|W|}$ is a standard technique in bounded-VC regularity~\cite{LS10}. Here we track progress across refinements via the Shannon entropy $H_W(\mu)$ of the cell weights themselves: holding cell masses fixed during the minimax step keeps $H_W$ invariant, while refining by just $2/\delta$ random draws from $p^*$ (rather than a $\wt{O}(d/\delta)$-point VC $\varepsilon$-net in every round, which would cost $\wt{O}(d^3)$) gains $\Omega(1)$ bits in expectation.

\section{Representative Cell Pool and Reconstruction via MWU}\label{sec:mwu}

In this section, we introduce the representative cell pool and show via the Multiplicative Weights Update (MWU) algorithm that once the compressor finds a small subset $W$ of the sample whose representative cell pool captures constant mass of a low-error distribution, reconstruction of the full sample follows directly. This reduces proving Theorem~\ref{thm:main} to constructing a small such set $W$.

By Ben-David and Litman's compactness lemma~\cite{BDL98} (see Lemma~\ref{lem:compactness} and Remark~\ref{rem:infinite}), if every finite restriction $\cC|_U$ ($U \subseteq \cX$ finite) admits a $(k,b)$-compression scheme with $(k,b)$ depending solely on $d$, then $\cC$ admits a $(k,b)$-scheme on $\cX$. Fixing $d \ge 1$, we therefore assume throughout Sections~\ref{sec:mwu}--\ref{sec:dichotomy} that $\cX$ is finite, so $\cC \subseteq \{0,1\}^{\cX}$ is finite and non-empty with $\VC(\cC) \le d$.

Before observing any sample, $\kappa$ and $\rho$ fix an arbitrary (strict) total order $\prec$ on $\cC$ (and an arbitrary total order on $\cX \times \{0,1\}$ for sorting subsets of $\cX \times \{0,1\}$).\footnote{Because $(\kappa, \rho)$ is defined for a fixed class $\cC$, fixing $\prec$ inside $\kappa$ and $\rho$ costs zero bits; cf.~\cite{FW95,MY16}.}

Given any non-empty\footnote{If $S$ is empty, $\kappa(S) := (\emptyset, \text{empty string})$ and $\rho(\emptyset, \sigma) := \min_\prec(\cC) \in \cC$ trivially suffice.} $\cC$-realizable sample $S = ((x_1, y_1), \dots, (x_m, y_m))$, let $X := \{x_1, \dots, x_m\} \subseteq \cX$ be the set of sample points, and let the compressor fix any concept $c^* \in \cC$ consistent with $S$ ($c^*(x_i) = y_i$ for all $i$). For any $c \in \cC$, $\mu \in \Delta(\cC)$, and $x \in X$, we slightly overload the notation and define the pointwise error of $c$ and of $\mu$ at $x$ by
\[
  \err_x(c) := \ind[c(x) \ne c^*(x)]
  \qquad \text{and} \qquad
  \err_x(\mu) := \Pr_{c \sim \mu}[c(x) \ne c^*(x)] = \sum_{c \in \cC} \mu(c)\,\err_x(c).
\]

\begin{definition}[Cells Induced by $W$ and the Representative Cell Pool]\label{def:pool}
Any subset $W \subseteq \cX$ partitions $\cC$ into equivalence classes, which we call the \emph{cells induced by $W$}, one for each realizable labeling $\tau \in \cC|_W$:
\[
  C_\tau := \{c \in \cC : c|_W = \tau\} \qquad (\tau \in \cC|_W).
\]
Let $\min_\prec(C_\tau) \in C_\tau$ denote the $\prec$-smallest concept (the \emph{representative}) of $C_\tau$. The \emph{representative cell pool} of $W$ is the set of all cell representatives:
\[
  \Pool(W) := \biggl\{\min_{\prec}(C_\tau) : \tau \in \cC|_W\biggr\} \subseteq \cC.
\]
\end{definition}

By the Sauer--Shelah--Perles lemma~\cite{VC71,Sau72,She72}, $|\Pool(W)| = |\cC|_W| \le \sum_{i=0}^{d} \binom{|W|}{i} \le (|W| + 1)^d$. Crucially, $\rho$ reconstructs $\Pool(W)$ from the \emph{unlabeled} set $W$ alone, using only $\cC$ and $\prec$.

The next result shows how to decode a distribution over $\Pool(W)$ from an unlabeled set $W$ whose representative cell pool supports a low-error mixture together with a short companion list of labeled points, yielding a weighted-majority voting rule that recovers the labels of the entire sample with zero error.

\begin{lemma}[MWU Reconstruction]\label{lem:mwu}
Let $W \subseteq X$ satisfy $|W| \le k_0$ ($k_0 \ge 1$), and suppose there exists a distribution $q \in \Delta(\Pool(W))$ with $\err_x(q) \le 1/4$ for all $x \in X$. Then, there exists a sequence of $t < k_1 := 8d\lceil \log_2(k_0 + 1) \rceil$ labeled points $(u_1, c^*(u_1)), \dots, (u_t, c^*(u_t))$ from $X$ (allowing repetitions) from which, together with $W$, the reconstructor recovers a hypothesis $h : \cX \to \{0,1\}$ satisfying $h(x) = c^*(x)$ for all $x \in X$.
\end{lemma}

\begin{proof}
We run the Multiplicative Weights Update (MWU) algorithm~\cite{LW94,FS97} on the expert set $\Pool(W)$ for $k_1$ rounds, answering in every round with a sample point that is the worst for its current strategy, and we use nothing about MWU beyond its worst-case regret bound (Fact~\ref{fact:mwu} in Appendix~\ref{app:classical}). Let $N := |\Pool(W)|$. In round $j = 1, \dots, k_1$, MWU plays a distribution $p_j \in \Delta(\Pool(W))$ that depends only on the losses revealed in rounds $1, \dots, j - 1$. We then pick a point
\[
  u_j \in \arg\max_{x \in X} \err_x(p_j)
\]
on which $p_j$ errs most, record $(u_j, c^*(u_j))$, and reveal to MWU the loss vector $\ell_j(c) := \err_{u_j}(c) = \ind[c(u_j) \ne c^*(u_j)] \in \{0,1\}$ (for all $c \in \Pool(W)$). The loss of MWU in round $j$ is therefore $\sum_{c} p_j(c)\,\ell_j(c) = \err_{u_j}(p_j) = \max_{x \in X} \err_x(p_j)$.

\emph{The average strategy is accurate at every point.} By Fact~\ref{fact:mwu}, the total loss of MWU exceeds that of the best fixed concept in $\Pool(W)$, and hence that of the mixture $q$, whose loss in every round is $\err_{u_j}(q) \le 1/4$, by at most $\sqrt{(k_1/2)\ln N}$:
\[
  \sum_{j=1}^{k_1} \err_{u_j}(p_j)
  \,\le\, \min_{c \in \Pool(W)} \sum_{j=1}^{k_1} \err_{u_j}(c) + \sqrt{\frac{k_1}{2}\ln N}
  \,\le\, \sum_{j=1}^{k_1} \err_{u_j}(q) + \sqrt{\frac{k_1}{2}\ln N}
  \,\le\, \frac{k_1}{4} + \sqrt{\frac{k_1}{2}\ln N}.
\]
Let $\bar{p} := \frac{1}{k_1}\sum_{j=1}^{k_1} p_j \in \Delta(\Pool(W))$ be the average strategy of MWU. Since $\mu \mapsto \err_x(\mu)$ is linear and $u_j$ is a worst point for $p_j$, every $x \in X$ satisfies
\[
  \err_x(\bar{p})
  \,=\, \frac{1}{k_1}\sum_{j=1}^{k_1} \err_x(p_j)
  \,\le\, \frac{1}{k_1}\sum_{j=1}^{k_1} \err_{u_j}(p_j)
  \,\le\, \frac{1}{4} + \sqrt{\frac{\ln N}{2 k_1}}
  \,<\, \frac{1}{4} + \frac{1}{4}
  \,=\, \frac{1}{2},
\]
where the strict inequality uses $N \le (k_0 + 1)^d \le 2^{k_1/8}$, so that $\ln N \le \frac{k_1}{8}\ln 2 < \frac{k_1}{8}$. Consequently, the weighted majority vote of $\bar{p}$,
\[
  h(x) := \ind\!\left[\sum_{c \in \Pool(W) :\, c(x) = 1} \bar{p}(c) > \frac{1}{2}\right],
\]
agrees with $c^*$ on all of $X$: at every $x \in X$, the $\bar{p}$-mass of the concepts voting against $c^*(x)$ satisfies $\err_x(\bar{p}) < \frac{1}{2}$.

\emph{Reconstruction.} Each loss vector $\ell_i$ is determined by the labeled point $(u_i, c^*(u_i))$, and $p_j$ is determined by $\ell_1, \dots, \ell_{j-1}$. Hence, given $W$ (which determines $\Pool(W)$ and $N$) and the $t := k_1 - 1$ labeled points $(u_1, c^*(u_1)), \dots, (u_t, c^*(u_t))$, the reconstructor replays MWU (with horizon $n = k_1 = t + 1$) to recompute $p_1, \dots, p_{k_1}$, and then $\bar{p}$ and $h$, without ever knowing $q$ or $X$. (The last point $u_{k_1}$ is needed only in the analysis.)
\end{proof}

\paragraph{Additional information for decoding.}
How does $\kappa$ convey both an unordered set $W \subseteq X$ ($|W| \le k_0$) and the MWU sequence $(u_1, c^*(u_1)), \dots, (u_t, c^*(u_t))$ ($0 \le t < k_1$) using a single unordered \emph{set} $T \subseteq \set(S) := \{(x_1, y_1), \dots, (x_m, y_m)\}$ and few side-information bits? Let $\kappa$ output the (deduplicated) union
\[
  T := \bigl\{(x, c^*(x)) : x \in W\bigr\} \cup \bigl\{(u_1, c^*(u_1)), \dots, (u_t, c^*(u_t))\bigr\} \subseteq \set(S),
\]
which has $|T| \le k := k_0 + k_1$, and let both $\kappa$ and $\rho$ sort $T$ under the fixed order on $\cX \times \{0,1\}$ as $z_1 < z_2 < \dots < z_{|T|}$. Because every element of $T$ that is \emph{not} one of the $t$ MWU points necessarily came from $W$, $\sigma$ only needs to describe those $t$ MWU steps: for each $j = 1, \dots, t$, write a fixed-width block of $B := \lceil \log_2 k \rceil + 1$ bits storing the index $i_j \in \{1, \dots, |T|\}$ where $(u_j, c^*(u_j)) = z_{i_j}$ together with one bit $\beta_j := \ind[u_j \in W] \in \{0,1\}$ indicating whether $u_j$ also belongs to $W$. This bitstring has total length $|\sigma| = t B \le k_1(\lceil \log_2 k \rceil + 1) =: b$. Given $(T, \sigma)$, the reconstructor divides $|\sigma|$ by $B$ to read $t$ and the full MWU sequence $(z_{i_1}, \dots, z_{i_t})$ (repeated indices $i_j = i_{j'}$ preserve any repeated MWU queries while occupying just one slot in $T$), and recovers $W$ as the $x$-coordinates of all unreferenced $z_i$ ($i \notin \{i_1, \dots, i_t\}$) together with those $z_{i_j}$ having $\beta_j = 1$.

\paragraph{Reduction to constructing a small set $W$.}
Finally, to apply Lemma~\ref{lem:mwu}, we do not necessarily need to construct a distribution supported solely on $\Pool(W)$: if we find \emph{any} distribution $\mu_{\mathrm{fin}} \in \Delta(\cC)$ and subset $W_{\mathrm{fin}} \subseteq X$ ($|W_{\mathrm{fin}}| \le k_0$, with $k_0 \ge 1$) satisfying
\begin{equation}\label{eq:exit}
  \err_x(\mu_{\mathrm{fin}}) \le \frac{1}{16} \quad (\forall x \in X)
  \qquad \text{and} \qquad
  \mu_{\mathrm{fin}}\bigl(\Pool(W_{\mathrm{fin}})\bigr) \ge \frac{1}{4},
\end{equation}
then conditioning $\mu_{\mathrm{fin}}$ on $\Pool(W_{\mathrm{fin}})$ yields $q(c) := \mu_{\mathrm{fin}}(c) / \mu_{\mathrm{fin}}(\Pool(W_{\mathrm{fin}})) \in \Delta(\Pool(W_{\mathrm{fin}}))$ with $\err_x(q) \le 4\,\err_x(\mu_{\mathrm{fin}}) \le 1/4$ for all $x \in X$. Applying Lemma~\ref{lem:mwu} and the decoding scheme above then immediately yields a sample compression scheme of size $k = k_0 + k_1 = k_0 + O(d \log(k_0 + 1))$ examples and $b = k_1(\lceil \log_2 k \rceil + 1) = O(d \log(k_0 + 1) \log(k_0 + d))$ side-information bits. Consequently, the remainder of the paper (Sections~\ref{sec:entropy}--\ref{sec:dichotomy}) is devoted to constructing a terminating pair $(W_{\mathrm{fin}}, \mu_{\mathrm{fin}})$ satisfying~\eqref{eq:exit} with $|W_{\mathrm{fin}}| \le k_0 = O(d^2 \log^2 d)$.

\section{Constructing a Small Set \texorpdfstring{$W$}{W} via Iterative Refinement}\label{sec:entropy}

Suppose we have a subset $W \subseteq X$ and a low-error distribution $\mu \in \Delta(\cC)$ (say $\err_x(\mu) \le 1/16$ on all $x \in X$). Why can't we satisfy~\eqref{eq:exit} immediately by moving the entire mass $\mu(C_\tau)$ of each cell $C_\tau$ induced by $W$ onto its representative $\min_\prec(C_\tau) \in \Pool(W)$? Because concepts in $C_\tau$ are only guaranteed to agree on $W$ and can behave arbitrarily on $X \setminus W$: high-error concepts in $C_\tau$ may precede the low-error ones under $\prec$, so moving the entire mass of $C_\tau$ onto $\min_\prec(C_\tau)$ could violate the pointwise error bound on $X$. Whenever this stops us from placing enough mass on $\min_\prec(C_\tau)$, we need to split $C_\tau$ into finer subcells to separate those earlier high-error concepts from the low-error ones after them; the natural way to refine the cells is to add new points from $X$ to $W$.

Our plan is, therefore, to find a terminating pair $(W_{\mathrm{fin}}, \mu_{\mathrm{fin}})$ obeying~\eqref{eq:exit} by iterating over rounds $r = 0, 1, 2, \dots$, and maintaining a subset $W_r \subseteq X$ and a low-error distribution $\mu_r \in \Delta(\cC)$. Fix a round limit $R = \Theta(d \log d)$ (chosen concretely in the proof of Theorem~\ref{thm:main} below) and error increment $\delta := 1/Q$ with $Q := 16 R$, so that $R\delta = 1/16$. At round $r = 0$, we initialize
\[
  W_0 := \emptyset
  \qquad \text{and} \qquad
  \mu_0 := \ind_{c^*} \,,
\]
the point mass on the consistent concept $c^*$, which satisfies $\err_x(\mu_0) = 0$ on all $x \in X$. In every non-terminating round $r$, we update $\mu_r$ to $\mu_{r+1}$ so as to shift mass inside the cells closer to their $\prec$-smallest representatives, at the cost of adding at most $2/\delta$ points to $W_r$ and allowing at most $\delta$ additional pointwise error on $X$. Thus, entering any round $r < R$, the pair $(W_r, \mu_r)$ obeys the two invariants
\[
  |W_r| \le \frac{2r}{\delta}
  \qquad \text{and} \qquad
  \err_x(\mu_r) \le r\delta \quad (\forall x \in X) \,.
\]
Since we iterate for fewer than $R$ rounds, the pointwise error $(r + 1)\delta \le R\delta = 1/16$ never exceeds our $1/16$ budget.

\paragraph{Tracking progress via cell entropy.}
To guarantee termination in fewer than $R$ rounds, we track a potential function built from the cell masses. For any $W \subseteq X$ and $\mu \in \Delta(\cC)$, let $w = (w_\tau)_{\tau \in \cC|_W}$ with $w_\tau := \mu(C_\tau)$ be the probability distribution that $\mu$ induces over the cells of $W$, and define its Shannon entropy
\begin{equation}\label{eq:entropy-bound}
  H_W(\mu) := \sum_{\tau \in \cC|_W :\, w_\tau > 0} w_\tau \log_2 \frac{1}{w_\tau}
  \,\le\, \log_2 |\cC|_W|
  \,\le\, d \log_2(|W| + 1),
\end{equation}
where the first inequality bounds the Shannon entropy of the discrete distribution $w$ by the logarithm of its support size, and the second follows from the Sauer--Shelah--Perles bound $|\cC|_W| \le (|W| + 1)^d$ (Definition~\ref{def:pool}).

At initialization, $W_0 = \emptyset$ induces only the single cell $C_\emptyset = \cC$ of mass $w_\emptyset = 1$, so $H_{W_0}(\mu_0) = 0$. We will show that in each round $r < R$, either we terminate (after adding one final batch of points to $W_r$ whose size is controlled) with a pair $(W_{\mathrm{fin}}, \mu_{\mathrm{fin}})$ satisfying~\eqref{eq:exit}, or we obtain $(W_{r+1}, \mu_{r+1})$ respecting our size and error invariants while increasing the potential $H_{W_r}(\mu_r)$ by an absolute constant ($> 1/128$ bits). Because $|W_r| \le 2r/\delta$ grows at most linearly in $r$, the upper bound $d \log_2(|W_r| + 1)$ in~\eqref{eq:entropy-bound} grows only logarithmically in $r$ whereas the entropy grows linearly in $r$. This guarantees termination in $O(d \log d)$ rounds. All remaining technical work is to establish this per-round invariant.

Because $H_W(\mu)$ depends on both $W$ and $\mu$, changing both at once would make it hard to track how $H_W(\mu)$ evolves. In Section~\ref{sec:dichotomy}, we therefore describe a single-round update procedure that splits into two stages: first, we hold $W_r$ fixed and shift mass leftward within each cell while preserving the cell entropy (Sections~\ref{subsec:qem} and~\ref{subsec:qem-proof}); second, we hold the resulting distribution fixed and enlarge $W_r$ to either terminate via an $\varepsilon$-net (Section~\ref{subsec:conc}) or gain more than $1/128$ bits of entropy via random draws (Section~\ref{subsec:spread}). The guarantee of this procedure is formalized in the following lemma.

\begin{lemma}[Single-Round Progress]\label{lem:dichotomy}
Let $\theta \ge 0$ and $\delta > 0$ satisfy $\theta + \delta \le 1/16$. Given any $W \subseteq X$ and $\mu \in \Delta(\cC)$ with $\err_x(\mu) \le \theta$ for all $x \in X$, there exists $W' \subseteq X$ containing $W$ and $\mu' \in \Delta(\cC)$ with $\err_x(\mu') \le \theta + \delta$ for all $x \in X$ such that at least one of the following holds:
\begin{enumerate}[label=(\arabic*), leftmargin=2.2em]
  \item \emph{(\textbf{Termination})} $|W' \setminus W| = O\bigl(\frac{d}{\delta}\log\frac{1}{\delta}\bigr)$ and $\mu'\bigl(\Pool(W')\bigr) \ge \frac{1}{4}$.
  \item \emph{(\textbf{Entropy Gain})} $|W' \setminus W| \le \lceil 2/\delta \rceil$ and $H_{W'}(\mu') > H_W(\mu) + \frac{1}{128}$.
\end{enumerate}
\end{lemma}

Combining Lemmas~\ref{lem:mwu} and~\ref{lem:dichotomy}, we can now prove Theorem~\ref{thm:main}.

\begin{proof}[Proof of Theorem~\ref{thm:main}]
Let $L := \lceil \log_2(d + 2) \rceil \ge 2$, $R := 4096\, d L = 2^{12} d L$, and $\delta := 1/(16 R)$, so that $R\delta = 1/16$ and $\lceil 2/\delta \rceil = 2/\delta = 32 R$. Initialize $W_0 := \emptyset$ and $\mu_0 := \ind_{c^*}$, which satisfy $H_{W_0}(\mu_0) = 0$ and $\err_x(\mu_0) = 0$ for all $x \in X$. For $r = 0, 1, \dots, R - 1$, we inductively maintain $\err_x(\mu_r) \le r\delta$ for all $x \in X$ and apply Lemma~\ref{lem:dichotomy} to $(W_r, \mu_r)$ with $\theta = r\delta$, noting that $\theta + \delta = (r + 1)\delta \le R\delta = 1/16$. If item~(1) holds, we terminate and set $(W_{\mathrm{fin}}, \mu_{\mathrm{fin}}) := (W', \mu')$; otherwise item~(2) holds, and we set $(W_{r+1}, \mu_{r+1}) := (W', \mu')$ and proceed to round $r + 1$.

Suppose towards contradiction that item~(2) of Lemma~\ref{lem:dichotomy} (Entropy Gain) held in all $R$ rounds $r = 0, \dots, R - 1$. By item~(2) of Lemma~\ref{lem:dichotomy}, each round increases the entropy potential by $H_{W_{r+1}}(\mu_{r+1}) - H_{W_r}(\mu_r) > 1/128$; summing from $H_{W_0}(\mu_0) = 0$ over $R$ rounds gives
\[
  H_{W_R}(\mu_R) > \frac{R}{128} = 32\, d L.
\]
Meanwhile, item~(2) of Lemma~\ref{lem:dichotomy} also ensures $|W_{r+1}| \le |W_r| + \lceil 2/\delta \rceil = |W_r| + 32 R$, so starting from $|W_0| = 0$ gives $|W_R| \le 32 R^2 = 2^{29} d^2 L^2$. Since $d < d + 2 \le 2^L$ and $L < 2^L$ (with $L \ge 2$), we have $|W_R| + 1 < 2^{29 + 4L} < 2^{32 L}$, and the entropy upper bound~\eqref{eq:entropy-bound} implied by the Sauer--Shelah--Perles lemma yields
\[
  H_{W_R}(\mu_R) \le d \log_2(|W_R| + 1) < 32\, d L,
\]
contradicting $H_{W_R}(\mu_R) > 32\, d L$.

Hence some round $r < R$ must satisfy item~(1) of Lemma~\ref{lem:dichotomy} (Terminate), after $r$ Entropy Gain rounds (so $|W_r| \le 32 r R < 32 R^2$), producing a terminating pair $(W_{\mathrm{fin}}, \mu_{\mathrm{fin}})$ obeying~\eqref{eq:exit} (since $\err_x(\mu_{\mathrm{fin}}) \le (r + 1)\delta \le R\delta = 1/16$) with
\[
  |W_{\mathrm{fin}}|
  \,\le\, |W_r| + C_0 \frac{d}{\delta}\log_2\frac{1}{\delta}
  \,\le\, \bigl\lceil 32 R^2 + 16 C_0\, d R \log_2(16 R) \bigr\rceil
  \,=:\, k_0 \,,
\]
where $C_0 \ge 1$ is the universal constant implicit in item~(1) of Lemma~\ref{lem:dichotomy} (so $k_0 \ge 1$ depends only on $d$). Conditioning $\mu_{\mathrm{fin}}$ on $\Pool(W_{\mathrm{fin}})$ as in~\eqref{eq:exit} and applying Lemma~\ref{lem:mwu} adds $t < k_1 := 8d\lceil \log_2(k_0 + 1) \rceil$ examples. Since $R = \Theta(d \log d)$ and $C_0$ is an absolute constant, we obtain $k_0 = O(d^2 \log^2 d)$, $k_1 = O(d \log d)$. Thus, the sample size is $k = k_0 + k_1 = O(d^2 \log^2 d)$ and we need $b = k_1(\lceil \log_2 k \rceil + 1) = O(d \log^2 d)$ side-information bits.
\end{proof}

\section{The Single-Round Progress}\label{sec:dichotomy}

Fix $W \subseteq X$, $\mu \in \Delta(\cC)$, $\theta \ge 0$, and $\delta > 0$ as in Lemma~\ref{lem:dichotomy}, and let $\theta' := \theta + \delta \le 1/16$. For each realizable labeling $\tau \in \cC|_W$, let $w_\tau := \mu(C_\tau)$ be the mass of the cell $C_\tau$ induced by $W$, let $\cS := \{\tau \in \cC|_W : w_\tau > 0\}$ index the active cells (so $\sum_{\tau \in \cS} w_\tau = 1$), and define $\cC_{\mathrm{act}} := \bigsqcup_{\tau \in \cS} C_\tau$. Note that $\cS$, $\cC_{\mathrm{act}}$, and $X$ are all finite non-empty sets. For any error threshold $t \ge 0$, define the feasible polytope
\[
  \cK(t) := \bigl\{\nu \in \Delta(\cC) : (\forall \tau \in \cS,\, \nu(C_\tau) = w_\tau) \;\text{ and }\; (\forall x \in X,\, \err_x(\nu) \le t)\bigr\}.
\]
Because any $\nu \in \cK(t)$ satisfies $\nu \ge 0$ and $\sum_{c \in \cC_{\mathrm{act}}} \nu(c) = \sum_{\tau \in \cS} w_\tau = 1 = \sum_{c \in \cC} \nu(c)$, we have $\nu(c) = 0$ for all $c \notin \cC_{\mathrm{act}}$; thus every $\nu \in \cK(t)$ is supported on $\cC_{\mathrm{act}}$ and induces the exact same cell masses $w_\tau$ as $\mu$, giving $H_W(\nu) = H_W(\mu)$. For each $\nu \in \cK(t)$ and $\tau \in \cS$, write $\nu_\tau(c) := \nu(c)/w_\tau$ for the conditional distribution of $\nu$ on $C_\tau$. Viewed in $[0,1]^{\cC_{\mathrm{act}}}$, $\cK(\theta')$ is defined by finitely many closed linear constraints and contains $\mu \in \cK(\theta) \subseteq \cK(\theta')$, so $\cK(\theta')$ is a non-empty and compact polytope.

It is instructive to think about $r = 0$; since $W_0 = \emptyset$, there is only a \emph{single} cell $C_\emptyset = \cC$ of mass $w_\emptyset = 1$ (so $\nu_\emptyset = \nu$), $\mu = \ind_{c^*} \in \cK(0)$, and $\cK(\delta)$ is simply the set of all distributions on $\cC$ with pointwise error $\le \delta$ on $X$. In a general round $r$, the exact same calculation runs across all active cells $C_\tau$ simultaneously, weighted by $w_\tau$. We keep $W, \mu, \theta, \theta', w_\tau, \cS$, and $\cK(t)$ fixed throughout this section. Section~\ref{subsec:qem} introduces the maximal lottery game over $\cK(\theta')$, states the resulting within-cell disagreement bound (Lemma~\ref{lem:qem}), and uses it to partition the active cells into concentrated and spread cells; Sections~\ref{subsec:conc} and~\ref{subsec:spread} handle those two cases; Section~\ref{subsec:dichotomy-proof} combines them to prove Lemma~\ref{lem:dichotomy}; and Section~\ref{subsec:qem-proof} proves Lemma~\ref{lem:qem} via linear programming duality.

\subsection{Maximal Lotteries}\label{subsec:qem}

Choosing a redistribution $\nu \in \cK(\theta')$ that pushes mass leftward requires care: although mass moves separately within each cell $C_\tau$, the error constraint $\err_x(\nu) = \sum_\tau \sum_{c \in C_\tau} \nu(c)\,\err_x(c) \le \theta'$ is \emph{global}, correlating all cells, so moving mass onto concepts that err at $x$ inside one cell uses up error budget at $x$ for all other cells. Natural greedy rules, such as pushing mass onto $\min_\prec(C_\tau)$ cell by cell, lack a clear way to measure progress.

Rather than scoring how far left a redistribution sits using some arbitrary numerical function of the positions under $\prec$, we compare two redistributions directly via the pairwise-comparison game of \emph{maximal lotteries} from probabilistic social choice~\cite{Kre65,Fis84} (see also~\cite{DHSSZ15}). Given $\nu, \eta \in \cK(\theta')$, sample an active cell $\tau \sim w$, draw $c_1 \sim \nu_\tau$ and $c_2 \sim \eta_\tau$ independently inside $C_\tau$, and declare the $\prec$-earlier of the two concepts the winner, breaking a tie $c_1 = c_2$ by a fair coin. Formally, for two concepts $c_1, c_2$ in a common cell, let
\[
  K(c_1, c_2) := \ind[c_2 \prec c_1] + \frac{1}{2}\,\ind[c_2 = c_1] \in \bigl\{0, \tfrac{1}{2}, 1\bigr\}
\]
denote the probability that $c_1$ loses to $c_2$, and let $M(\nu, \eta)$ be the probability that $\nu$ loses to $\eta$:
\begin{align*}
  M(\nu, \eta)
  &:= \sum_{\tau \in \cS} w_\tau \sum_{c_1, c_2 \in C_\tau} \nu_\tau(c_1)\,\eta_\tau(c_2)\,K(c_1, c_2) \\
  &\phantom{:}= \sum_{\tau \in \cS} \frac{1}{w_\tau} \sum_{c_1, c_2 \in C_\tau} \nu(c_1)\,\eta(c_2)\,K(c_1, c_2),
\end{align*}
where the second equality uses $\nu_\tau(c) = \nu(c)/w_\tau$ and $\eta_\tau(c) = \eta(c)/w_\tau$ on each active cell $C_\tau$ (recalling that $w_\tau > 0$ is fixed across $\cK(\theta')$). Moving mass to an earlier concept of the same cell never increases a player's loss (if $c' \prec c$, then $K(c', c_2) \le K(c, c_2)$ for every $c_2$). To find a redistribution that no feasible opponent beats head-to-head, we consider the constrained min--max problem
\begin{equation}\label{eq:minmax}
  \min_{\nu \in \cK(\theta')} \ \max_{\eta \in \cK(\theta')} \ M(\nu, \eta),
\end{equation}
in which the $\nu$-player minimizes $M(\nu, \eta)$ over $\cK(\theta')$ and the $\eta$-player maximizes it over the same polytope $\cK(\theta')$. Note that $M(\nu, \eta)$ is bilinear on $\cK(\theta') \times \cK(\theta')$, and since $\prec$ is a strict total order, $K(c_1, c_2) + K(c_2, c_1) = 1$ for every $c_1, c_2 \in C_\tau$. Summing against $\nu$ and $\eta$ gives the constant-sum identity
\begin{equation}\label{eq:sym}
  M(\nu, \eta) + M(\eta, \nu) \,=\, \sum_{\tau \in \cS} w_\tau \,=\, 1
  \qquad \text{for all } \nu, \eta \in \cK(\theta'),
\end{equation}
and in particular $M(\eta, \eta) = \frac{1}{2}$ for all $\eta \in \cK(\theta')$.

\begin{claim}[Existence of a Maximal Lottery]\label{clm:saddle}
There exists $\mu^* \in \cK(\theta')$, called a \emph{maximal lottery} over $\cK(\theta')$, such that $(\mu^*, \mu^*)$ is a symmetric saddle point of~\eqref{eq:minmax}:
\begin{equation}\label{eq:saddle}
  M(\mu^*, \eta) \,\le\, \frac{1}{2} \,=\, M(\mu^*, \mu^*) \,\le\, M(\nu, \mu^*)
  \qquad \text{for all } \nu, \eta \in \cK(\theta').
\end{equation}
\end{claim}

\begin{proof}
Since $M$ is bilinear and $\cK(\theta')$ is a non-empty compact convex polytope, von Neumann's minimax theorem (Fact~\ref{fact:lp}(a)) provides a saddle point $(\hat{\nu}, \hat{\eta}) \in \cK(\theta') \times \cK(\theta')$ of~\eqref{eq:minmax},\footnote{In the notation of Fact~\ref{fact:lp}: the variables are indexed by $\cC_{\mathrm{act}}$; $\cK(\theta') = \{\nu \in [0, \infty)^{\cC_{\mathrm{act}}} : A\nu = w,\, E\nu \le \theta'\ind\}$ with $A_{\tau, c} := \ind[c \in C_\tau]$ and $E_{x, c} := \err_x(c)$; and $M(\nu, \eta) = \nu^\top Q \eta$ with $Q_{c_1, c_2} := K(c_1, c_2)/w_\tau$ if $c_1, c_2$ lie in a common cell $C_\tau$ and $Q_{c_1, c_2} := 0$ otherwise.} i.e., $M(\hat{\nu}, \eta) \le M(\hat{\nu}, \hat{\eta}) \le M(\nu, \hat{\eta})$ for all $\nu, \eta \in \cK(\theta')$. Evaluating the left inequality at $\eta = \hat{\nu}$ and the right inequality at $\nu = \hat{\eta}$ traps the game value at $\frac{1}{2} = M(\hat{\nu}, \hat{\nu}) \le M(\hat{\nu}, \hat{\eta}) \le M(\hat{\eta}, \hat{\eta}) = \frac{1}{2}$. Hence $\mu^* := \hat{\eta} \in \cK(\theta')$ satisfies $M(\nu, \mu^*) \ge \frac{1}{2}$ for all $\nu \in \cK(\theta')$, which is the right half of~\eqref{eq:saddle}, and~\eqref{eq:sym} gives the left half: $M(\mu^*, \eta) = 1 - M(\eta, \mu^*) \le \frac{1}{2}$ for all $\eta \in \cK(\theta')$.
\end{proof}

Fix a maximal lottery $\mu^* \in \cK(\theta')$ given by Claim~\ref{clm:saddle}. By~\eqref{eq:sym}, the two halves of~\eqref{eq:saddle} are equivalent, so $(\mu^*, \mu^*)$ being a saddle point is captured by the single optimality condition that $\mu^*$ is a best response to itself:
\begin{equation}\label{eq:lp}
  \mu^* \in \arg\min_{\nu \in \cK(\theta')} M(\nu, \mu^*).
\end{equation}
For each active cell $\tau \in \cS$ and concept $c \in C_\tau$, define the \emph{within-cell quantile} of $c$ under $\mu^*$ as the probability that $c$ loses to a random draw from $\mu^*_\tau$:
\begin{equation}\label{eq:quantile}
  r_\tau(c) \,:=\, \sum_{c' \in C_\tau} \mu^*_\tau(c')\,K(c, c') \,=\, \mu^*_\tau\bigl(\{c' \in C_\tau : c' \prec c\}\bigr) + \frac{1}{2}\,\mu^*_\tau(c) \in [0, 1].
\end{equation}
Note that for any distribution $\mu^* \in \cK(\theta')$ and its within-cell quantiles~\eqref{eq:quantile}, whenever $c_1, c_2 \in C_\tau$ satisfy $c_1 \prec c_2$, every $c' \preceq c_1$ strictly precedes $c_2$, so $\{c' \in C_\tau : c' \preceq c_1\} \subseteq \{c' \in C_\tau : c' \prec c_2\}$ and therefore
\begin{equation}\label{eq:mono}
  r_\tau(c_1)
  \,\le\, \mu^*_\tau\bigl(\{c' \in C_\tau : c' \preceq c_1\}\bigr)
  \,\le\, \mu^*_\tau\bigl(\{c' \in C_\tau : c' \prec c_2\}\bigr)
  \,=\, r_\tau(c_2) - \frac{1}{2}\,\mu^*_\tau(c_2).
\end{equation}
Factoring $\nu(c)$ out of $M(\nu, \mu^*)$ expresses the objective in~\eqref{eq:lp} as a linear functional of $\nu$ with coefficients $r_\tau(c)$:
\begin{equation}\label{eq:linear-obj}
  M(\nu, \mu^*)
  \,=\, \sum_{\tau \in \cS} \sum_{c \in C_\tau} \nu(c) \sum_{c' \in C_\tau} \mu^*_\tau(c')\,K(c, c')
  \,=\, \sum_{\tau \in \cS} \sum_{c \in C_\tau} \nu(c)\, r_\tau(c).
\end{equation}
Shifting mass inside $C_\tau$ from a positive-mass concept $c_2$ to an earlier concept $c_1 \prec c_2$ preserves every cell mass $w_\tau$ while decreasing~\eqref{eq:linear-obj} by $r_\tau(c_2) - r_\tau(c_1) > 0$ per unit of shifted mass, so $\mu^*$ can refrain from doing so only when blocked by tight error constraints $\err_x(\mu^*) = \theta'$. Applying linear programming duality to~\eqref{eq:lp} and exploiting the uniform $\delta$-slack of $\mu \in \cK(\theta)$\footnote{Recall $\mu$ is the distribution that the current round started with.} yields a single distribution $p^* \in \Delta(X)$ satisfying the following disagreement lower bound.

\begin{lemma}[Disagreement Bound for Maximal Lotteries]\label{lem:qem}
Let $\mu^* \in \cK(\theta')$ be a maximal lottery (Claim~\ref{clm:saddle}), with within-cell quantiles $(r_\tau)_{\tau \in \cS}$ defined by~\eqref{eq:quantile}. Then at least one of the following holds:
\begin{enumerate}[label=\textup{(\arabic*)}, leftmargin=2.2em]
  \item $\mu^*\bigl(\Pool(W)\bigr) = 1$; or
  \item there exists a distribution $p^* \in \Delta(X)$ such that, for every $\tau \in \cS$ and every $c_1, c_2 \in C_\tau$ with $c_1 \prec c_2$ and $\mu^*(c_2) > 0$,
  \begin{equation}\label{eq:qem}
    d_{p^*}(c_1, c_2) \,\ge\, 2\delta\bigl(r_\tau(c_2) - r_\tau(c_1)\bigr),
  \end{equation}
  where $d_{p^*}(c_1, c_2) := \Pr_{x \sim p^*}[c_1(x) \ne c_2(x)]$. In particular, any $c_1, c_2 \in C_\tau$ with $\mu^*(c_1) > 0$ and $\mu^*(c_2) > 0$ satisfy $d_{p^*}(c_1, c_2) \ge 2\delta\,|r_\tau(c_1) - r_\tau(c_2)|$.
\end{enumerate}
\end{lemma}

We defer the proof of Lemma~\ref{lem:qem} to Section~\ref{subsec:qem-proof} and first show how it implies Lemma~\ref{lem:dichotomy}. Notice the useful asymmetry of~\eqref{eq:qem}: only the later concept $c_2$ needs positive mass $\mu^*(c_2) > 0$, while the earlier concept $c_1 \prec c_2$ is allowed to carry zero mass, i.e., $\mu^*(c_1) = 0$.

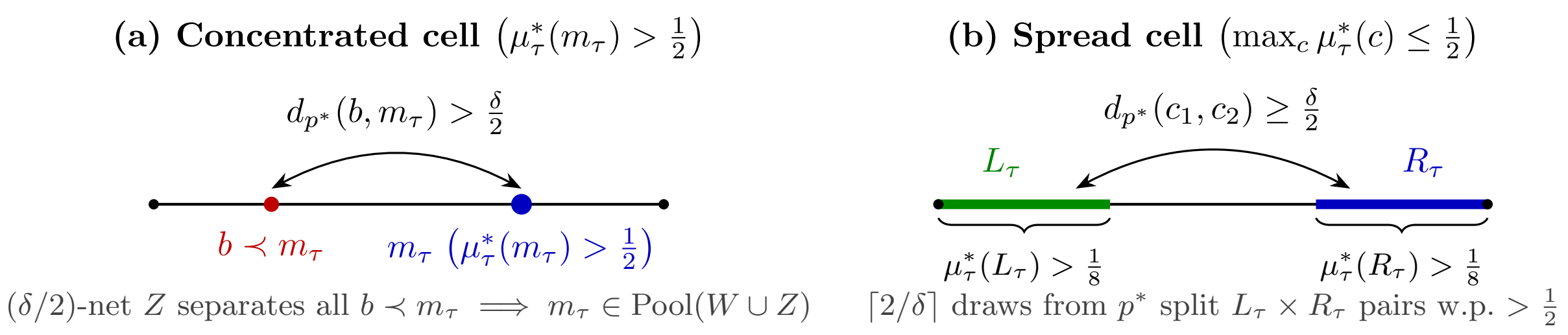
\begin{figure}[t]
\centering
\begin{tikzpicture}[>=Stealth, font=\small]
  \begin{scope}[xshift=0cm]
    \node[font=\small\bfseries] at (2.6, 1.7) {(a) Concentrated cell $\bigl(\mu^*_\tau(m_\tau) > \tfrac{1}{2}\bigr)$};
    \draw[thick] (0,0) -- (5.2,0);
    \fill (0,0) circle (1.5pt);
    \fill (5.2,0) circle (1.5pt);

    \fill[red!75!black] (1.2,0) circle (2.2pt);
    \node[red!75!black, below=4pt] at (1.2,0) {$b \prec m_\tau$};

    \fill[blue!75!black] (3.75,0) circle (3.0pt);
    \node[blue!75!black, below=3pt] at (3.75,0) {$m_\tau \;\bigl(\mu^*_\tau(m_\tau) > \tfrac{1}{2}\bigr)$};

    \draw[<->, semithick] (1.2, 0.16) to[bend left=28] node[above=2pt] {$d_{p^*}(b, m_\tau) > \tfrac{\delta}{2}$} (3.75, 0.16);

    \node[black!75, font=\footnotesize] at (2.6, -1.05) {$(\delta/2)$-net $Z$ separates all $b \prec m_\tau \implies m_\tau \in \Pool(W \cup Z)$};
  \end{scope}

  \begin{scope}[xshift=8.0cm]
    \node[font=\small\bfseries] at (2.8, 1.7) {(b) Spread cell $\bigl(\max_c \mu^*_\tau(c) \le \tfrac{1}{2}\bigr)$};
    \draw[thick] (0,0) -- (5.6,0);
    \draw[green!55!black, line width=2.8pt] (0,0) -- (1.75,0);
    \draw[blue!75!black, line width=2.8pt] (3.85,0) -- (5.6,0);
    \fill (0,0) circle (1.5pt);
    \fill (5.6,0) circle (1.5pt);

    \node[green!50!black, above=3pt] at (0.65, 0.04) {$L_\tau$};
    \node[blue!75!black, above=3pt] at (4.95, 0.04) {$R_\tau$};

    \draw[decorate, decoration={brace, mirror, amplitude=4pt}, semithick] (0, -0.14) -- (1.75, -0.14)
      node[midway, below=5pt, font=\footnotesize] {$\mu^*_\tau(L_\tau) > \tfrac{1}{8}$};
    \draw[decorate, decoration={brace, mirror, amplitude=4pt}, semithick] (3.85, -0.14) -- (5.6, -0.14)
      node[midway, below=5pt, font=\footnotesize] {$\mu^*_\tau(R_\tau) > \tfrac{1}{8}$};

    \draw[<->, semithick] (1.4, 0.16) to[bend left=28] node[above=2pt] {$d_{p^*}(c_1, c_2) \ge \tfrac{\delta}{2}$} (4.2, 0.16);

    \node[black!75, font=\footnotesize] at (2.8, -1.05) {$\lceil 2/\delta \rceil$ draws from $p^*$ split $L_\tau \times R_\tau$ pairs w.p.\ $> \tfrac{1}{2}$};
  \end{scope}
\end{tikzpicture}
\caption{The two types of active cells $C_\tau$ under the disagreement bound~\eqref{eq:qem}. In (a), $r_\tau(m_\tau) - r_\tau(b) \ge \frac{1}{2}\mu^*_\tau(m_\tau) > \frac{1}{4}$ forces $d_{p^*}(b, m_\tau) > \frac{\delta}{2}$ for \emph{every} earlier concept $b \prec m_\tau$. In (b), $L_\tau = \{r_\tau \le \frac{3}{8}\}$ and $R_\tau = \{r_\tau \ge \frac{5}{8}\}$ each carry conditional mass $> \frac{1}{8}$ and lie at $d_{p^*}$-distance $\ge \frac{\delta}{2}$, so $\lceil 2/\delta \rceil$ random draws from $p^*$ split each fixed positive-mass pair with probability $> \frac{1}{2}$.}
\label{fig:cells}
\end{figure}

\paragraph{Concentrated cells and spread cells.}
Assuming item~(2) of Lemma~\ref{lem:qem} holds (otherwise item~(1) already gives $\mu^*(\Pool(W)) = 1$ and immediate termination), \eqref{eq:qem} turns positive quantile distance inside any active cell $C_\tau$ into disagreement probability under $p^*$: whenever $c_1 \prec c_2$ in $C_\tau$ satisfy $\mu^*(c_2) > 0$ and $r_\tau(c_2) - r_\tau(c_1) \ge \frac{1}{4}$, they obey $d_{p^*}(c_1, c_2) \ge \frac{\delta}{2}$. To see where such quantile gaps arise, view $\mu^*_\tau$ as laying out the concepts of $C_\tau$ in $\prec$-order along the unit interval $[0, 1]$, where each concept $c$ occupies a sub-interval of length $\mu^*_\tau(c)$ whose midpoint is $r_\tau(c)$. This geometry splits the active cells $\cS$ into two classes according to whether some single concept carries more than half of $\mu^*_\tau$ (Figure~\ref{fig:cells}):
\[
  \cS_{\mathrm{conc}} := \Bigl\{\tau \in \cS : \max_{c \in C_\tau} \mu^*_\tau(c) > \frac{1}{2}\Bigr\},
  \qquad
  \cS_{\mathrm{spread}} := \Bigl\{\tau \in \cS : \max_{c \in C_\tau} \mu^*_\tau(c) \le \frac{1}{2}\Bigr\}.
\]
In a \emph{concentrated} cell $\tau \in \cS_{\mathrm{conc}}$, a single concept $m_\tau$ occupies an interval of length $> \frac{1}{2}$; the atom at $m_\tau$ cannot itself be split by a refinement, but its own half-width places $r_\tau(m_\tau)$ at distance $> \frac{1}{4}$ above \emph{every} earlier concept $b \prec m_\tau$, so the asymmetry of~\eqref{eq:qem} gives $d_{p^*}(b, m_\tau) > \frac{\delta}{2}$ for all $b \prec m_\tau$ simultaneously and a single $(\delta/2)$-net makes $m_\tau$ the representative of its refined cell. Conversely, in a \emph{spread} cell $\tau \in \cS_{\mathrm{spread}}$, no single concept occupies more than half of $[0, 1]$, so the mass must straddle both sides of the middle quantile band $(\frac{3}{8}, \frac{5}{8})$, leaving mass $> \frac{1}{8}$ on $\{r_\tau \le \frac{3}{8}\}$ and $> \frac{1}{8}$ on $\{r_\tau \ge \frac{5}{8}\}$; every positive-mass pair across these two sides has quantile gap $\ge \frac{1}{4}$, hence $d_{p^*} \ge \frac{\delta}{2}$, so $\lceil 2/\delta \rceil$ random draws from $p^*$ split those pairs and raise $H_W(\mu^*)$. Since $\sum_{\tau \in \cS_{\mathrm{conc}}} w_\tau + \sum_{\tau \in \cS_{\mathrm{spread}}} w_\tau = 1$, either $\sum_{\tau \in \cS_{\mathrm{conc}}} w_\tau \ge \frac{1}{2}$ (handled in Section~\ref{subsec:conc}) or $\sum_{\tau \in \cS_{\mathrm{spread}}} w_\tau > \frac{1}{2}$ (handled in Section~\ref{subsec:spread}); we combine these cases to prove Lemma~\ref{lem:dichotomy} in Section~\ref{subsec:dichotomy-proof}.

\subsection{Concentrated Cells: Termination via an \texorpdfstring{$\varepsilon$}{epsilon}-Net}\label{subsec:conc}

We first handle the case in which concentrated cells carry at least half the total cell weight.

\begin{lemma}[Termination from Concentrated Cells]\label{lem:conc}
Let $\mu^* \in \cK(\theta')$ and $p^* \in \Delta(X)$ satisfy item~\textup{(2)} of Lemma~\ref{lem:qem}, with within-cell quantiles $(r_\tau)_{\tau \in \cS}$ as in~\eqref{eq:quantile} (so~\eqref{eq:mono} and~\eqref{eq:qem} hold). Suppose $\sum_{\tau \in \cS_{\mathrm{conc}}} w_\tau \ge \frac{1}{2}$. Then, there exists $Z \subseteq X$ with $|Z| \le 25 C_{\mathrm{net}} \frac{d}{\delta}\log_2\frac{1}{\delta}$ (where $C_{\mathrm{net}} \ge 1$ is the constant of Fact~\ref{fact:eps-net}) such that
\[
  \mu^*\bigl(\Pool(W \cup Z)\bigr) \,>\, \frac{1}{4}.
\]
\end{lemma}

\begin{proof}
For each $\tau \in \cS_{\mathrm{conc}}$, since $\sum_{c \in C_\tau} \mu^*_\tau(c) = 1$, there is a unique concept $m_\tau \in C_\tau$ satisfying $\mu^*_\tau(m_\tau) > \frac{1}{2}$ (hence $\mu^*(m_\tau) = w_\tau \mu^*_\tau(m_\tau) > \frac{1}{2} w_\tau > 0$). For every earlier concept $b \in C_\tau$ with $b \prec m_\tau$ (if any exist; when $m_\tau = \min_\prec(C_\tau)$ there are none), combining~\eqref{eq:mono} and~\eqref{eq:qem} gives
\[
  d_{p^*}(b, m_\tau)
  \,\ge\, 2\delta\bigl(r_\tau(m_\tau) - r_\tau(b)\bigr)
  \,\ge\, \delta\,\mu^*_\tau(m_\tau)
  \,>\, \frac{\delta}{2}.
\]

To separate every $m_\tau$ from all such $b \prec m_\tau$ simultaneously, define the symmetric difference class $\cC \oplus \cC := \{c_1 \oplus c_2 : c_1, c_2 \in \cC\}$, where $(c_1 \oplus c_2)(x) := \ind[c_1(x) \ne c_2(x)]$. On any finite subset $A \subseteq \cX$ of size $s \ge d$, Sauer's lemma implies $|\cC|_A| \le \sum_{i=0}^d \binom{s}{i} \le (es/d)^d$, so $|(\cC \oplus \cC)|_A| \le |\cC|_A|^2 \le (es/d)^{2d}$. Whenever $\alpha := s/d \ge 10$, we have $2^{\alpha/2} > e\alpha$ (at $\alpha = 10$, $32 > 10e$, and the left side grows faster for $\alpha \ge 10$), hence $(es/d)^{2d} < 2^s$; thus no subset of size $s \ge 10d$ is shattered, giving $\VC\bigl((\cC \oplus \cC)|_X\bigr) \le \VC(\cC \oplus \cC) < 10d$. Applying Fact~\ref{fact:eps-net} (Appendix~\ref{app:classical}) to $\cF = (\cC \oplus \cC)|_X$ with $v = 10d$, distribution $p^*$, and $\varepsilon = \delta/2 \le \theta'/2 \le 1/32$ produces a subset $Z \subseteq X$ of size
\[
  |Z|
  \,\le\, C_{\mathrm{net}} \frac{10d}{\delta/2} \log_2 \frac{2}{\delta}
  \,=\, 20 C_{\mathrm{net}} \frac{d}{\delta}\Bigl(1 + \log_2 \frac{1}{\delta}\Bigr)
  \,\le\, 25 C_{\mathrm{net}} \frac{d}{\delta} \log_2 \frac{1}{\delta}
\]
(using $\log_2(1/\delta) \ge \log_2 16 = 4$) such that $d_{p^*}(c_1, c_2) \ge \delta/2 \implies c_1|_Z \ne c_2|_Z$.

Now take any $\tau \in \cS_{\mathrm{conc}}$ and any concept $c \in \cC$ satisfying $c|_{W \cup Z} = m_\tau|_{W \cup Z}$. Restricting to $W$ shows $c|_W = m_\tau|_W = \tau$, so $c \in C_\tau$. Moreover, we cannot have $c \prec m_\tau$, because every $b \in C_\tau$ with $b \prec m_\tau$ satisfies $d_{p^*}(b, m_\tau) > \delta/2$ and therefore $b|_Z \ne m_\tau|_Z$. Since $\prec$ is a total order, $m_\tau \preceq c$ holds for all $c$ in $m_\tau$'s refined cell, meaning $m_\tau = \min_\prec\{c \in \cC : c|_{W \cup Z} = m_\tau|_{W \cup Z}\} \in \Pool(W \cup Z)$ (including when $m_\tau = \min_\prec(C_\tau)$). Because the cells $(C_\tau)_{\tau \in \cS_{\mathrm{conc}}}$ are pairwise disjoint, the concepts $(m_\tau)_{\tau \in \cS_{\mathrm{conc}}}$ are distinct elements of $\Pool(W \cup Z)$, giving
\[
  \mu^*\bigl(\Pool(W \cup Z)\bigr)
  \,\ge\, \sum_{\tau \in \cS_{\mathrm{conc}}} \mu^*(m_\tau)
  \,>\, \sum_{\tau \in \cS_{\mathrm{conc}}} \frac{1}{2}\, w_\tau
  \,\ge\, \frac{1}{4}. \qedhere
\]
\end{proof}

\subsection{Spread Cells: Constant Entropy Gain via Random Draws}\label{subsec:spread}

We first mention an elementary lower bound showing that refining $W$ to $W' = W \cup S_0$ increases the cell Shannon entropy of any fixed distribution $\nu \in \cK(\theta')$ by at least the within-cell non-collision probability.

\begin{lemma}[Collision Lower Bound on Entropy Gain]\label{lem:collision}
For any subset $S_0 \subseteq X$, setting $W' := W \cup S_0$, every distribution $\nu \in \cK(\theta')$ satisfies
\[
  H_{W'}(\nu) - H_W(\nu)
  \,\ge\, \sum_{\tau \in \cS} w_\tau \sum_{c_1, c_2 \in C_\tau} \nu_\tau(c_1)\,\nu_\tau(c_2)\,\ind\bigl[c_1|_{W'} \ne c_2|_{W'}\bigr].
\]
\end{lemma}

\begin{proof}
Each active cell $C_\tau$ ($\tau \in \cS$) is partitioned into the finer cells $\{C_{\tau'} : \tau' \in \cC|_{W'},\, \tau'|_W = \tau\}$ induced by $W'$. Let $q_{\tau, \tau'} := \nu_\tau(C_{\tau'}) = \sum_{c \in C_{\tau'}} \nu_\tau(c)$ be the conditional share of $C_\tau$'s mass falling into $C_{\tau'}$, so $\nu(C_{\tau'}) = w_\tau q_{\tau, \tau'}$ and $\sum_{\tau'} q_{\tau, \tau'} = 1$. Expanding $\log_2 \frac{1}{w_\tau q_{\tau, \tau'}} = \log_2 \frac{1}{w_\tau} + \log_2 \frac{1}{q_{\tau, \tau'}}$ in $H_{W'}(\nu)$, and using $\log_2(1/q) = \frac{1}{\ln 2}\ln(1/q) \ge \ln(1/q) \ge 1 - q$ for all $q \in (0, 1]$, gives
\begin{align*}
  H_{W'}(\nu) - H_W(\nu)
  &= \sum_{\tau \in \cS} \sum_{\tau' :\, q_{\tau, \tau'} > 0} w_\tau q_{\tau, \tau'} \Bigl(\log_2 \frac{1}{w_\tau} + \log_2 \frac{1}{q_{\tau, \tau'}}\Bigr) - \sum_{\tau \in \cS} w_\tau \log_2 \frac{1}{w_\tau} \\
  &= \sum_{\tau \in \cS} w_\tau \sum_{\tau' :\, q_{\tau, \tau'} > 0} q_{\tau, \tau'} \log_2 \frac{1}{q_{\tau, \tau'}}
  \,\ge\, \sum_{\tau \in \cS} w_\tau \Bigl(1 - \sum_{\tau'} q_{\tau, \tau'}^2\Bigr) \\
  &= \sum_{\tau \in \cS} w_\tau \sum_{c_1, c_2 \in C_\tau} \nu_\tau(c_1)\,\nu_\tau(c_2)\,\ind\bigl[c_1|_{W'} \ne c_2|_{W'}\bigr],
\end{align*}
since $\sum_{\tau'} q_{\tau, \tau'}^2 = \sum_{\tau'} \bigl(\sum_{c \in C_{\tau'}} \nu_\tau(c)\bigr)^2 = \sum_{c_1, c_2 \in C_\tau} \nu_\tau(c_1)\,\nu_\tau(c_2)\,\ind[c_1|_{W'} = c_2|_{W'}]$ and the product weights $\nu_\tau(c_1)\,\nu_\tau(c_2)$ sum to $1$ over $C_\tau \times C_\tau$.
\end{proof}

We now use the previous fact to prove the per-round entropy gaini from spread cells.

\begin{lemma}[Entropy Gain from Spread Cells]\label{lem:spread}
Let $\mu^* \in \cK(\theta')$ and $p^* \in \Delta(X)$ satisfy item~\textup{(2)} of Lemma~\ref{lem:qem}, with within-cell quantiles $(r_\tau)_{\tau \in \cS}$ as in~\eqref{eq:quantile} (so~\eqref{eq:mono} and~\eqref{eq:qem} hold). Suppose $\sum_{\tau \in \cS_{\mathrm{spread}}} w_\tau > \frac{1}{2}$. Then, there exists $S^* \subseteq X$ with $|S^*| \le \lceil 2/\delta \rceil$ such that
\[
  H_{W \cup S^*}(\mu^*) - H_W(\mu^*) \,>\, \frac{1}{128}.
\]
\end{lemma}

\begin{proof}
Fix any spread cell $\tau \in \cS_{\mathrm{spread}}$ (so $\mu^*_\tau(c) \le \frac{1}{2}$ for all $c \in C_\tau$), and define its left and right quantile blocks
\[
  L_\tau := \Bigl\{c \in C_\tau : r_\tau(c) \le \frac{3}{8}\Bigr\}
  \qquad \text{and} \qquad
  R_\tau := \Bigl\{c \in C_\tau : r_\tau(c) \ge \frac{5}{8}\Bigr\}.
\]
Clearly $L_\tau \cap R_\tau = \emptyset$. Let us verify that $\mu^*_\tau(L_\tau) > \frac{1}{8}$ and $\mu^*_\tau(R_\tau) > \frac{1}{8}$:
\begin{itemize}[leftmargin=1.5em]
  \item Let $c_{\max} := \max_\prec(C_\tau)$; then $r_\tau(c_{\max}) = (1 - \mu^*_\tau(c_{\max})) + \frac{1}{2}\mu^*_\tau(c_{\max}) = 1 - \frac{1}{2}\mu^*_\tau(c_{\max}) \ge \frac{3}{4} > \frac{3}{8}$, so $C_\tau \setminus L_\tau$ is non-empty. Let $c^- := \min_\prec(C_\tau \setminus L_\tau)$. Every $c \in C_\tau$ with $c \prec c^-$ lies in $L_\tau$ by $\prec$-minimality of $c^-$, while every $c \in C_\tau$ with $c^- \preceq c$ obeys $r_\tau(c) \ge r_\tau(c^-) > \frac{3}{8}$ by~\eqref{eq:mono}; hence $L_\tau = \{c \in C_\tau : c \prec c^-\}$. Using $\mu^*_\tau(c^-) \le \frac{1}{2}$ then gives
  \[
    \frac{3}{8} < r_\tau(c^-) = \mu^*_\tau(L_\tau) + \frac{1}{2}\,\mu^*_\tau(c^-) \le \mu^*_\tau(L_\tau) + \frac{1}{4}
    \quad \implies \quad
    \mu^*_\tau(L_\tau) > \frac{1}{8}.
  \]
  \item Symmetrically, let $c_{\min} := \min_\prec(C_\tau)$; then $r_\tau(c_{\min}) = \frac{1}{2}\mu^*_\tau(c_{\min}) \le \frac{1}{4} < \frac{5}{8}$, so $C_\tau \setminus R_\tau$ is non-empty. Let $c^+ := \max_\prec(C_\tau \setminus R_\tau)$; by $\prec$-maximality of $c^+$ and monotonicity~\eqref{eq:mono}, we have $R_\tau = \{c \in C_\tau : c^+ \prec c\}$, so $\{c \in C_\tau : c \prec c^+\}$ has conditional mass $1 - \mu^*_\tau(R_\tau) - \mu^*_\tau(c^+)$. Using $\mu^*_\tau(c^+) \le \frac{1}{2}$ gives
  \[
    \frac{5}{8} > r_\tau(c^+) = 1 - \mu^*_\tau(R_\tau) - \frac{1}{2}\,\mu^*_\tau(c^+) \ge \frac{3}{4} - \mu^*_\tau(R_\tau)
    \quad \implies \quad
    \mu^*_\tau(R_\tau) > \frac{1}{8}.
  \]
\end{itemize}
Moreover, for any $c_1 \in L_\tau$ and $c_2 \in R_\tau$ with $\mu^*(c_1), \mu^*(c_2) > 0$, we have $|r_\tau(c_1) - r_\tau(c_2)| \ge \frac{5}{8} - \frac{3}{8} = \frac{1}{4}$, so~\eqref{eq:qem} yields $d_{p^*}(c_1, c_2) \ge \frac{\delta}{2}$ (and identically when $c_1 \in R_\tau, c_2 \in L_\tau$).

Finally, we apply the probabilistic method to show that some subset of size $\le \lceil 2/\delta \rceil$ makes the lower bound of Lemma~\ref{lem:collision} exceed $1/128$. Set $\ell := \lceil 2/\delta \rceil$, draw $\ell$ points $z_1, \dots, z_\ell \stackrel{\mathrm{i.i.d.}}{\sim} p^*$, and form the random subset $S_{\mathrm{new}} := \{z_1, \dots, z_\ell\} \subseteq X$ (note $|S_{\mathrm{new}}| \le \ell$ for every outcome). For any \emph{fixed} pair $c_1, c_2 \in C_\tau$ with $d_{p^*}(c_1, c_2) \ge \frac{\delta}{2}$, we have $c_1|_{W \cup S_{\mathrm{new}}} = c_2|_{W \cup S_{\mathrm{new}}}$ only when all $\ell$ draws miss $\{x \in X : c_1(x) \ne c_2(x)\}$; using $1 - u \le e^{-u}$ and $\ell \ge 2/\delta$ gives
\[
  \Pr_{S_{\mathrm{new}}}\bigl[c_1|_{W \cup S_{\mathrm{new}}} \ne c_2|_{W \cup S_{\mathrm{new}}}\bigr]
  \,\ge\, 1 - \bigl(1 - d_{p^*}(c_1, c_2)\bigr)^{\ell}
  \,\ge\, 1 - (1 - \delta/2)^{\ell}
  \,\ge\, 1 - e^{-\ell\delta/2}
  \,\ge\, 1 - e^{-1}
  \,>\, \frac{1}{2}.
\]
Applying Lemma~\ref{lem:collision} to $\nu = \mu^*$ and $S_0 = S_{\mathrm{new}}$ and taking expectations over $S_{\mathrm{new}}$ via linearity of expectation over the finite sum gives
\begin{align*}
  \mathbb{E}_{S_{\mathrm{new}}}\bigl[H_{W \cup S_{\mathrm{new}}}(\mu^*) - H_W(\mu^*)\bigr]
  &\ge \sum_{\tau \in \cS} w_\tau \sum_{c_1, c_2 \in C_\tau} \mu^*_\tau(c_1)\,\mu^*_\tau(c_2)\,\Pr_{S_{\mathrm{new}}}\bigl[c_1|_{W \cup S_{\mathrm{new}}} \ne c_2|_{W \cup S_{\mathrm{new}}}\bigr] \\
  &\ge \sum_{\tau \in \cS_{\mathrm{spread}}} w_\tau \sum_{(c_1, c_2) \in (L_\tau \times R_\tau) \sqcup (R_\tau \times L_\tau)} \mu^*_\tau(c_1)\,\mu^*_\tau(c_2) \cdot \frac{1}{2} \\
  &= \sum_{\tau \in \cS_{\mathrm{spread}}} w_\tau \cdot 2\,\mu^*_\tau(L_\tau)\,\mu^*_\tau(R_\tau) \cdot \frac{1}{2}
  \,>\, \frac{1}{2} \cdot \frac{1}{8} \cdot \frac{1}{8}
  \,=\, \frac{1}{128},
\end{align*}
where the second inequality drops non-negative terms outside $\cS_{\mathrm{spread}}$ and outside $(L_\tau \times R_\tau) \sqcup (R_\tau \times L_\tau)$ (whenever $\mu^*_\tau(c_1)\,\mu^*_\tau(c_2) > 0$ on those blocks, we verified $d_{p^*}(c_1, c_2) \ge \delta/2$ and hence split probability $> 1/2$). Because the expectation strictly exceeds $1/128$, there exists a deterministic outcome $S^* \subseteq X$ in the support of $S_{\mathrm{new}}$ (so $|S^*| \le \ell = \lceil 2/\delta \rceil$) whose realized entropy satisfies $H_{W \cup S^*}(\mu^*) - H_W(\mu^*) > 1/128$.
\end{proof}

\subsection{Proof of \texorpdfstring{Lemma~\ref{lem:dichotomy}}{the Single-Round Dichotomy}}\label{subsec:dichotomy-proof}

Combining Claim~\ref{clm:saddle} and Lemmas~\ref{lem:qem}, \ref{lem:conc}, and~\ref{lem:spread} now proves Lemma~\ref{lem:dichotomy}.

\begin{proof}[Proof of Lemma~\ref{lem:dichotomy}]
Let $\mu^* \in \cK(\theta')$ be a maximal lottery given by Claim~\ref{clm:saddle}, with within-cell quantiles $(r_\tau)_{\tau \in \cS}$ as in~\eqref{eq:quantile}. In every case below we set $\mu' := \mu^*$, which satisfies $\err_x(\mu') \le \theta' = \theta + \delta$ for all $x \in X$ and $H_W(\mu') = H_W(\mu)$ because $\mu^* \in \cK(\theta')$.

If item~(1) of Lemma~\ref{lem:qem} holds, then $\mu^*(\Pool(W)) = 1 \ge \frac{1}{4}$, so setting $W' := W$ immediately gives item~(1) (\textbf{Termination}) with $|W' \setminus W| = 0$.

Otherwise, item~(2) of Lemma~\ref{lem:qem} provides a distribution $p^* \in \Delta(X)$ satisfying~\eqref{eq:qem}. Since $\sum_{\tau \in \cS_{\mathrm{conc}}} w_\tau + \sum_{\tau \in \cS_{\mathrm{spread}}} w_\tau = 1$, one of two cases holds:
\begin{itemize}[leftmargin=1.5em]
  \item \emph{Concentrated cells dominate ($\sum_{\tau \in \cS_{\mathrm{conc}}} w_\tau \ge \frac{1}{2}$).}
  Lemma~\ref{lem:conc} yields $Z \subseteq X$ with
  \[
    |Z| \,\le\, 25 C_{\mathrm{net}} \frac{d}{\delta}\log_2\frac{1}{\delta}
    \,=\, O\Bigl(\frac{d}{\delta}\log\frac{1}{\delta}\Bigr)
    \qquad \text{and} \qquad
    \mu^*\bigl(\Pool(W \cup Z)\bigr) \,>\, \frac{1}{4},
  \]
  where the asymptotic bound uses $\delta \le \theta' \le 1/16$. Setting $W' := W \cup Z$ gives item~(1) (\textbf{Termination}), since $|W' \setminus W| \le |Z|$.

  \item \emph{Spread cells dominate ($\sum_{\tau \in \cS_{\mathrm{spread}}} w_\tau > \frac{1}{2}$).}
  Lemma~\ref{lem:spread} yields $S^* \subseteq X$ with $|S^*| \le \lceil 2/\delta \rceil$ and
  \[
    H_{W \cup S^*}(\mu^*) \,>\, H_W(\mu^*) + \frac{1}{128} \,=\, H_W(\mu) + \frac{1}{128}.
  \]
  Setting $W' := W \cup S^*$ gives item~(2) (\textbf{Entropy Gain}), since $|W' \setminus W| \le |S^*| \le \lceil 2/\delta \rceil$. \qedhere
\end{itemize}
\end{proof}

\subsection{Proof of \texorpdfstring{Lemma~\ref{lem:qem}}{the Disagreement Bound} via LP Duality}\label{subsec:qem-proof}

Let $\mu^* \in \cK(\theta')$ be a maximal lottery (Claim~\ref{clm:saddle}) with within-cell quantiles $(r_\tau)_{\tau \in \cS}$ as defined in~\eqref{eq:quantile}. By~\eqref{eq:lp} and~\eqref{eq:linear-obj}, $\mu^*$ minimizes the linear functional $\nu \mapsto M(\nu, \mu^*) = \sum_{\tau \in \cS} \sum_{c \in C_\tau} \nu(c)\, r_\tau(c)$ over $\cK(\theta')$, attaining minimum value $M(\mu^*, \mu^*) = \frac{1}{2}$. Applying linear programming duality to~\eqref{eq:lp}, the next claim extracts non-negative dual multipliers $\lambda_x \ge 0$ with $\ell_1$-norm $\Lambda := \sum_{x \in X} \lambda_x \le \frac{1}{2\delta}$; we then normalize $(\lambda_x)_{x \in X}$ into $p^*$ when $\Lambda > 0$ (or conclude $\mu^*(\Pool(W)) = 1$ when $\Lambda = 0$).

\begin{claim}[Bounded Dual Multipliers]\label{clm:prices}
Let $\mu^* \in \cK(\theta')$ be a maximal lottery (Claim~\ref{clm:saddle}) with within-cell quantiles $(r_\tau)_{\tau \in \cS}$ as in~\eqref{eq:quantile}. There exist non-negative dual multipliers $\lambda_x \ge 0$ ($x \in X$) satisfying
\[
  \Lambda := \sum_{x \in X} \lambda_x \,\le\, \frac{1}{2\delta}
\]
such that, for every active cell $\tau \in \cS$ and every $c_1, c_2 \in C_\tau$ with $\mu^*(c_2) > 0$,
\begin{equation}\label{eq:cs-diff}
  \sum_{x \in X} \lambda_x \bigl(\err_x(c_1) - \err_x(c_2)\bigr) \,\ge\, r_\tau(c_2) - r_\tau(c_1).
\end{equation}
\end{claim}

\begin{proof}
\emph{Step 1: Dual multipliers via LP duality (normal cone).}
View $\cK(\theta')$ as a polytope in $\mathbb{R}^{\cC_{\mathrm{act}}}$, and let $r := (r_\tau(c))_{c \in \cC_{\mathrm{act}}}$ be the gradient of the linear function $\nu \mapsto M(\nu, \mu^*)$. Then~\eqref{eq:lp} shows that $\langle r, \nu - \mu^* \rangle \ge 0$ for every $\nu \in \cK(\theta')$; in other words, the negative gradient lies in the normal cone of $\cK(\theta')$ at $\mu^*$:
\begin{equation}\label{eq:normal}
  -r \in N(\mu^*) := \bigl\{v \in \mathbb{R}^{\cC_{\mathrm{act}}} : \langle v, \nu - \mu^* \rangle \le 0 \ \text{ for all } \nu \in \cK(\theta')\bigr\}.
\end{equation}
The polytope $\cK(\theta')$ is cut out by three families of linear constraints, and the normal cone of a polytope at one of its points is generated by the outward normals of the constraints that hold with equality at that point (Fact~\ref{fact:lp}(b)). At $\mu^*$ these are:
\begin{itemize}[leftmargin=1.5em]
  \item every cell-mass equation $\sum_{c \in C_\tau} \nu(c) = w_\tau$ ($\tau \in \cS$); its normal is the indicator vector $\ind_{C_\tau}$ of the cell, and since it is an equation it may enter with a coefficient of either sign;
  \item the error constraint $\sum_{c \in \cC_{\mathrm{act}}} \err_x(c)\,\nu(c) \le \theta'$ at each tight point, i.e., each $x \in X$ with $\err_x(\mu^*) = \theta'$; its outward normal is the vector $\err_x := (\err_x(c))_{c \in \cC_{\mathrm{act}}}$;
  \item the sign constraint $\nu(c) \ge 0$ of each concept with $\mu^*(c) = 0$; its outward normal is $-e_c$, the negative of the $c$-th coordinate vector.
\end{itemize}
Hence
\[
  N(\mu^*) \,=\, \biggl\{-\sum_{\tau \in \cS} a_\tau \ind_{C_\tau} + \sum_{x :\, \err_x(\mu^*) = \theta'} \lambda_x \err_x - \sum_{c :\, \mu^*(c) = 0} s_c\, e_c \;:\; a_\tau \in \mathbb{R},\; \lambda_x \ge 0,\; s_c \ge 0\biggr\},
\]
and~\eqref{eq:normal} shows that $-r$ can be written in this form. Let $\lambda_x := 0$ at the points that are not tight and $s_c := 0$ at the concepts with $\mu^*(c) > 0$, and compare the coordinates of a concept $c \in C_\tau$ on the two sides:
\[
  -r_\tau(c) \,=\, -a_\tau + \sum_{x \in X} \lambda_x \err_x(c) - s_c.
\]
Since $s_c \ge 0$, with $s_c = 0$ whenever $\mu^*(c) > 0$, this is equivalent to
\begin{equation}\label{eq:kkt}
  r_\tau(c) + \sum_{x \in X} \lambda_x \err_x(c) \ge a_\tau
  \quad (\forall \tau \in \cS,\, \forall c \in C_\tau),
  \qquad \text{with equality whenever } \mu^*(c) > 0,
\end{equation}
where $\lambda_x \ge 0$ for all $x \in X$ and $\lambda_x > 0$ only if $\err_x(\mu^*) = \theta'$. In the language of linear programming, $a_\tau$, $\lambda_x$, and $s_c$ are the multipliers of the three families of constraints, and the fact that only the constraints active at $\mu^*$ contribute is complementary slackness. If $c_1, c_2 \in C_\tau$ and $\mu^*(c_2) > 0$, then subtracting the equality for $c_2$ from the inequality for $c_1$ in~\eqref{eq:kkt} cancels $a_\tau$ and yields~\eqref{eq:cs-diff}. Viewing $\sum_{x \in X} \lambda_x \err_x(c)$ as the Lagrangian penalty for errors at the tight points,~\eqref{eq:kkt} shows that inside each cell $C_\tau$ every concept has penalized quantile at least $a_\tau$ and $\mu^*$ places mass only on minimizers, while~\eqref{eq:cs-diff} shows that any quantile advantage $r_\tau(c_2) - r_\tau(c_1)$ of $c_1$ over a positive-mass concept $c_2$ is offset by higher $\lambda$-weighted error.

\emph{Step 2: Slater bound on $\Lambda$.}
Before normalizing $(\lambda_x)_{x \in X}$ into a distribution on $X$, we upper-bound $\Lambda := \sum_{x \in X} \lambda_x$ so that dividing~\eqref{eq:cs-diff} by $\Lambda$ preserves a quantitative lower bound. Multiply~\eqref{eq:kkt} by $\nu(c) \ge 0$ and sum over $c \in \cC_{\mathrm{act}}$: since $\nu(C_\tau) = w_\tau$ and $\sum_{\tau \in \cS}\sum_{c \in C_\tau} \nu(c)\, r_\tau(c) = M(\nu, \mu^*)$, every $\nu \in \cK(\theta')$ satisfies $M(\nu, \mu^*) + \sum_{x \in X} \lambda_x \err_x(\nu) \ge \sum_{\tau \in \cS} w_\tau a_\tau$, with equality at $\nu = \mu^*$. Subtracting $\theta'\Lambda$ from both sides, and noting that $\sum_{x \in X} \lambda_x\bigl(\err_x(\mu^*) - \theta'\bigr) = 0$ because $\lambda_x > 0$ only when $\err_x(\mu^*) = \theta'$, we obtain the Lagrangian inequality
\begin{equation}\label{eq:priced}
  M(\nu, \mu^*) + \sum_{x \in X} \lambda_x\bigl(\err_x(\nu) - \theta'\bigr) \,\ge\, M(\mu^*, \mu^*) \,=\, \frac{1}{2}
  \qquad \text{for every } \nu \in \cK(\theta').
\end{equation}
Evaluating~\eqref{eq:priced} at the incoming distribution $\mu \in \cK(\theta) \subseteq \cK(\theta')$, which satisfies the stricter error bound $\err_x(\mu) \le \theta = \theta' - \delta$ at every $x \in X$ (and using $\lambda_x \ge 0$ and $M(\mu, \mu^*) \le 1$), gives
\[
  \frac{1}{2} \,\le\, M(\mu, \mu^*) - \Lambda\delta \,\le\, 1 - \Lambda\delta,
  \qquad \text{and hence} \qquad
  \Lambda \,\le\, \frac{1}{2\delta},
\]
which is the classical Slater bound on the dual multipliers noted in Section~\ref{sec:related}.
\end{proof}

\paragraph{Example.}
To see the game and its optimality condition at work, take a single cell (as in round $r = 0$) that consists of just two concepts $b \prec c^*$, where $b$ errs at a single sample point $x_0 \in X$ and $c^*$ errs nowhere. A feasible $\nu$ is determined by $\nu(b) \in [0, \theta']$, and evaluating the definition gives $M(\nu, \eta) = \frac{1}{2} + \frac{1}{2}\bigl(\eta(b) - \nu(b)\bigr)$: each player simply wants as much mass on the earlier concept $b$ as the error constraint at $x_0$ permits. The symmetric saddle point is therefore $\mu^*(b) = \theta'$ and $\mu^*(c^*) = 1 - \theta'$, with quantiles $r(b) = \frac{\theta'}{2}$ and $r(c^*) = \frac{1 + \theta'}{2}$. Both concepts carry mass, so the optimality condition~\eqref{eq:kkt} from Step~1 of the proof of Claim~\ref{clm:prices} holds with equality at both, $\frac{\theta'}{2} + \lambda_{x_0} = a = \frac{1 + \theta'}{2}$ (where $a$ is the single-cell multiplier), while $\lambda_x = 0$ at every other point because only $x_0$ is tight. Hence $\lambda_{x_0} = \frac{1}{2} = r(c^*) - r(b)$: the multiplier $\lambda_{x_0}$ at $x_0$ offsets the quantile advantage of $b$ over $c^*$, which is why $\mu^*$ keeps mass on both concepts.

Claim~\ref{clm:prices} now yields Lemma~\ref{lem:qem}.

\begin{proof}[Proof of Lemma~\ref{lem:qem}]
Let $(\lambda_x)_{x \in X}$ and $\Lambda := \sum_{x \in X} \lambda_x \in [0, \frac{1}{2\delta}]$ be the dual multipliers given by Claim~\ref{clm:prices} for the maximal lottery $\mu^* \in \cK(\theta')$.

\emph{Degenerate case $\Lambda = 0$.}
If $\Lambda = 0$, then $\lambda_x = 0$ for all $x \in X$ (since $\lambda_x \ge 0$). Fix any active cell $\tau \in \cS$ and let $c_1 := \min_\prec(C_\tau) \in \Pool(W)$. For every $c_2 \in C_\tau$ with $\mu^*(c_2) > 0$, \eqref{eq:cs-diff} gives $r_\tau(c_2) \le r_\tau(c_1)$, which forces $c_2 = c_1$ because $c_1 \prec c_2$ would yield $r_\tau(c_1) \le r_\tau(c_2) - \frac{1}{2}\mu^*_\tau(c_2) < r_\tau(c_2)$ by~\eqref{eq:mono}. Thus $\mu^*$ is supported entirely on $\Pool(W)$, so $\mu^*(\Pool(W)) = \sum_{\tau \in \cS} w_\tau = 1$, establishing item~(1).

\emph{Main case $0 < \Lambda \le \frac{1}{2\delta}$.}
Normalize $p^*(x) := \lambda_x / \Lambda \in \Delta(X)$. For each $x \in X$, the difference $\err_x(c_1) - \err_x(c_2) \in \{-1, 0, 1\}$ equals $0$ whenever $c_1(x) = c_2(x)$ and is at most $1$ whenever $c_1(x) \ne c_2(x)$, so $\err_x(c_1) - \err_x(c_2) \le \ind[c_1(x) \ne c_2(x)]$. Dividing~\eqref{eq:cs-diff} by $\Lambda \in (0, \frac{1}{2\delta}]$, and using $r_\tau(c_2) - r_\tau(c_1) \ge 0$ when $c_1 \prec c_2$ (by~\eqref{eq:mono}) so that $\frac{1}{\Lambda} \ge 2\delta$ preserves the inequality direction, gives
\[
  d_{p^*}(c_1, c_2)
  \,\ge\, \frac{r_\tau(c_2) - r_\tau(c_1)}{\Lambda}
  \,\ge\, 2\delta\bigl(r_\tau(c_2) - r_\tau(c_1)\bigr),
\]
establishing~\eqref{eq:qem} and hence item~(2). The symmetric consequence for $\mu^*(c_1), \mu^*(c_2) > 0$ follows immediately from symmetry of $d_{p^*}$ and equality when $c_1 = c_2$. (In the Example above, $\Lambda = \frac{1}{2}$, $p^*$ is the point mass at $x_0$, and the first inequality above holds with equality for $c_1 = b$ and $c_2 = c^*$: $d_{p^*}(b, c^*) = 1 = (r(c^*) - r(b))/\Lambda$.)
\end{proof}

\section{Conclusion}
In this work we have shown that every VC class admits a sample compression scheme of size $\wt{O}(d^2)$,  giving the first improvement over the exponential bound of Moran and Yehudayoff~\cite{MY16}. This resolves the weaker version of the sample compression conjecture, which asks for a bound that is polynomial in $d$. Interestingly, the main overhead in our compression scheme comes from the \emph{unlabeled} points. We hope that the ideas presented here can pave the path towards a scheme of size $O(d)$, which would resolve the sample compression conjecture affirmatively.

\section*{AI Disclosure}
\phantomsection\label{sec:ai-disclosure}
The proof was obtained autonomously by the Cogentic multi-agent orchestration, whose description can be found in~\cite{CGJLMVW26}, using Gemini~4 Carbon. The agent was prompted to resolve the sample compression conjecture, without being given any hints or ideas of approaches that might work. To be more precise, it was instructed to obtain results that fall into three different tiers: (Tier 1) full resolution of the conjecture, perhaps up to polylogarithmic factors, (Tier 2) a polynomial bound on the sample compression size, and (Tier 3) a subexponential bound on the sample compression size.
The total cost, based on the current Gemini API pricing, was approximately \$344.
The Cogentic agent initially came up with an $\wt{O}(d^{16})$ scheme, then improved it to $\wt{O}(d^6)$ and $\wt{O}(d^3)$, subsequently to $O(d^2 \log^3 d)$, and lastly it gave the bound $O(d^2 \log^2 d)$ that we present in this work.
 The authors have subsequently worked to simplify the proof and improve the exposition, as well as to point out connections of the techniques used in the proof to prior work. The authors take responsibility for the correctness of the results and the accuracy of the exposition.

\newpage
\bibliographystyle{plain}
\bibliography{references}

\newpage

\appendix
\section{Technical Lemmas}\label{app:classical}

For completeness, we state the VC $\varepsilon$-net theorem invoked once in Section~\ref{subsec:conc}, include the short Tychonoff compactness argument reducing infinite domains to finite ones in Section~\ref{sec:mwu} (together with a direct well-ordering alternative), and record the two classical facts from online learning and optimization used as black boxes in Lemma~\ref{lem:mwu} and Sections~\ref{subsec:qem} and~\ref{subsec:qem-proof}.

\begin{fact}[VC $\varepsilon$-Net Theorem~\cite{HW87}]\label{fact:eps-net}
There exists an absolute constant $C_{\mathrm{net}} \ge 1$ such that for every non-empty finite set $X$, every family $\cF \subseteq \{0,1\}^X$ with $\VC(\cF) \le v$ ($v \ge 1$), every distribution $p \in \Delta(X)$, and every $\varepsilon \in (0, 1/2]$, there exists a subset $Z \subseteq X$ of size $|Z| \le C_{\mathrm{net}} \frac{v}{\varepsilon}\log_2\frac{1}{\varepsilon}$ (an \emph{$\varepsilon$-net}) such that every $f \in \cF$ with $\Pr_{x \sim p}[f(x) = 1] \ge \varepsilon$ satisfies $f(z) = 1$ for some $z \in Z$.
\end{fact}

\begin{lemma}[Compactness Lift to Infinite Domains~\cite{BDL98}]\label{lem:compactness}
Let $k, b \in \mathbb{Z}_{\ge 0}$ and $\cC \subseteq \{0,1\}^{\cX}$. If for every finite subset $U \subseteq \cX$ the restricted class $\cC|_U \subseteq \{0,1\}^U$ admits a labeled sample compression scheme of size $(k, b)$, then $\cC$ admits a labeled sample compression scheme of size $(k, b)$ on $\cX$.
\end{lemma}

\begin{proof}
Let $\Sigma := \{0,1\}^{\le b}$ (which is finite) and let $\cM := \{T \subseteq \cX \times \{0,1\} : |T| \le k\}$. Endow $\{0,1\}$ with the discrete topology; by Tychonoff's theorem, the product space $\Omega := \{0,1\}^{\cM \times \Sigma \times \cX}$ of all candidate reconstructors $\rho : \cM \times \Sigma \to \{0,1\}^{\cX}$ is compact. For each finite $\cC$-realizable sample $S = ((x_1, y_1), \dots, (x_m, y_m))$, define
\[
  F_S := \Bigl\{\rho \in \Omega : \exists\, T \subseteq \set(S) \text{ with } |T| \le k,\; \exists\, \sigma \in \Sigma \text{ s.t. } \forall i \in [m],\; \rho(T, \sigma)(x_i) = y_i\Bigr\}.
\]
Since $\set(S)$ and $\Sigma$ are finite, membership in $F_S$ depends on only finitely many coordinates $(T, \sigma, x_i) \in \cM \times \Sigma \times \cX$, so each $F_S$ is closed in $\Omega$. Moreover, for any finite collection of finite $\cC$-realizable samples $S^{(1)}, \dots, S^{(n)}$, let $U \subseteq \cX$ be the finite set of all points appearing in them; extending a $(k, b)$-reconstructor $\rho_U$ for $\cC|_U$ arbitrarily outside $U$ yields an element of $\bigcap_{\ell=1}^n F_{S^{(\ell)}} \ne \emptyset$. By compactness of $\Omega$, the full intersection $\bigcap_S F_S$ is non-empty; fix any $\rho \in \bigcap_S F_S$, and for each finite $\cC$-realizable sample $S$ let $\kappa(S) := (T, \sigma)$ be any pair witnessing $\rho \in F_S$.
\end{proof}

\begin{remark}[Infinite Domains via Well-Ordering]\label{rem:infinite}
Under the Axiom of Choice one can avoid Tychonoff compactness altogether and retain an explicit weighted-majority vote over proper concepts in $\cC$ on infinite domains $\cX$: fix a global well-ordering $\prec$ of $\cC$ (and a linear order on $\cX \times \{0,1\}$). For any finite $W \subseteq \cX$, $\Pool(W) = \{\min_\prec(C_\tau) : \tau \in \cC|_W\} \subseteq \cC$ is finite of size $\le (|W| + 1)^d$. On any finite sample domain $X \subseteq \cX$, restricting to the finite class $\Pool(X) \subseteq \cC$ (which is in bijection with $\cC|_X$) runs the finite-domain proof of Sections~\ref{sec:mwu}--\ref{sec:dichotomy} verbatim because $\Pool(W) \subseteq \Pool(X)$ holds for every $W \subseteq X$: indeed, for any $c_\tau = \min_\prec(C_\tau) \in \Pool(W)$, the representative $c' \in \Pool(X)$ matching $c_\tau$ on $X$ also matches $c_\tau$ on $W \subseteq X$, so $\prec$-minimality in both cells gives $c' \preceq c_\tau \preceq c'$, i.e., $c_\tau = c' \in \Pool(X)$.
\end{remark}

\begin{fact}[Regret Bound of Multiplicative Weights~{\cite[Theorem~2.2 and Corollary~2.2]{CBL06}}; see also~{\cite{LW94,FS97}}]\label{fact:mwu}
Let $\cP$ be a non-empty finite set of $N$ experts, let $n \ge 1$ be a number of rounds, and set $\eta := \sqrt{8 \ln N / n}$. In each round $j = 1, \dots, n$, the Multiplicative Weights Update (MWU) algorithm plays the distribution $p_j \in \Delta(\cP)$ given by $p_j(c) \propto \exp\bigl(-\eta \sum_{i < j} \ell_i(c)\bigr)$ and then observes a loss vector $\ell_j \in [0, 1]^{\cP}$; in particular, $p_j$ is a function of $\ell_1, \dots, \ell_{j-1}$ only. For every sequence of loss vectors, even one in which each $\ell_j$ is chosen adaptively after seeing $p_1, \dots, p_j$, it holds that
\[
  \sum_{j=1}^n \sum_{c \in \cP} p_j(c)\,\ell_j(c)
  \,\le\, \min_{c \in \cP} \sum_{j=1}^n \ell_j(c) + \sqrt{\frac{n}{2}\ln N}.
\]
\end{fact}


\begin{fact}[Minimax Theorem and Normal Cones of Polytopes~{\cite{vN28,Sch86}}]\label{fact:lp}
Let $\cK = \{\nu \in [0, \infty)^n : A\nu = w,\, E\nu \le g\}$ be non-empty and bounded, where $A \in \mathbb{R}^{m_1 \times n}$, $w \in \mathbb{R}^{m_1}$, $E \in \mathbb{R}^{m_2 \times n}$, and $g \in \mathbb{R}^{m_2}$.
\begin{enumerate}[label=\textup{(\alph*)}, leftmargin=2.2em]
  \item \textup{(von Neumann's minimax theorem.)} For every $Q \in \mathbb{R}^{n \times n}$, the bilinear function $M(\nu, \eta) := \nu^\top Q \eta$ has a saddle point on $\cK \times \cK$: there exist $\hat{\nu}, \hat{\eta} \in \cK$ with $M(\hat{\nu}, \eta) \le M(\hat{\nu}, \hat{\eta}) \le M(\nu, \hat{\eta})$ for all $\nu, \eta \in \cK$.
  \item \textup{(Normal cone of a polytope.)} For $\nu^* \in \cK$, the normal cone $N_{\cK}(\nu^*) := \{v \in \mathbb{R}^n : \langle v, \nu - \nu^* \rangle \le 0 \text{ for all } \nu \in \cK\}$ is generated by the outward normals of the constraints of $\cK$ that hold with equality at $\nu^*$: it consists of the vectors $A^\top \alpha + E^\top \lambda - s$ with $\alpha \in \mathbb{R}^{m_1}$, with $\lambda \in [0, \infty)^{m_2}$ such that $\lambda_x = 0$ for every row $x$ with $(E\nu^*)_x < g_x$, and with $s \in [0, \infty)^n$ such that $s_c = 0$ for every coordinate $c$ with $\nu^*_c > 0$.
\end{enumerate}
\end{fact}

Both parts of Fact~\ref{fact:lp} are textbook consequences of linear-programming duality. Part~(a) is von Neumann's minimax theorem~\cite{vN28} for the matrix game whose pure strategies are the finitely many vertices of $\cK$: every point of $\cK$ is a convex combination of vertices and $M$ is bilinear, so a pair of optimal mixed strategies over the vertices yields a saddle point in $\cK \times \cK$. Part~(b) is Farkas' lemma~\cite[Corollary~7.1d]{Sch86}: every vector of the stated form lies in $N_{\cK}(\nu^*)$, because for $\nu \in \cK$ we have $\langle A^\top \alpha + E^\top \lambda - s,\, \nu - \nu^* \rangle = \lambda^\top(E\nu - g) - s^\top \nu \le 0$; conversely, if $v$ is not of this form, Farkas' lemma provides a direction $d$ with $Ad = 0$, with $(Ed)_x \le 0$ at the rows where $(E\nu^*)_x = g_x$, with $d_c \ge 0$ at the coordinates where $\nu^*_c = 0$, and with $\langle v, d \rangle > 0$, whence $\nu^* + \varepsilon d \in \cK$ for all sufficiently small $\varepsilon > 0$ while $\langle v, (\nu^* + \varepsilon d) - \nu^* \rangle > 0$, so $v \notin N_{\cK}(\nu^*)$.

\section{Well-Spread Low-Error Distributions Need Not Admit a Small Split}\label{app:vc1-example}

In Section~\ref{sec:overview}, at round $t$, we seek a small set of unlabeled examples from the sample that refines the current partition $\cP_t$ so that either the cell representatives capture mass at least $1/4$ or the cell-mass entropy increases by $\Omega(1)$. Here we give an explicit example of VC dimension $d = 1$ (at $t = 0$, where $\cP_0 = \{\cC\}$ consists of a single cell of mass $1$) showing that a well-spread arbitrary distribution $\mu$ in the feasible polytope, even one that is uniform over arbitrarily many concepts, need not admit any bounded-size split achieving either goal, whereas choosing $\mu$ as a maximal lottery circumvents this obstruction.

Recall from Definition~\ref{def:pool} and~\eqref{eq:entropy-bound} that for a finite class $\cC \subseteq \{0,1\}^{\cX}$, a strict linear order $\prec$ on $\cC$, and a subset $W \subseteq \cX$, writing $c|_W : W \to \{0,1\}$ for the restriction of $c$ to $W$ and $\cC|_W := \{c|_W : c \in \cC\}$, the cells induced by $W$ are $C_\tau := \{c \in \cC : c|_W = \tau\}$ ($\tau \in \cC|_W$), the set of cell representatives is $\Pool(W) := \{\min_\prec(C_\tau) : \tau \in \cC|_W\}$, and for any distribution $\mu \in \Delta(\cC)$ (with $\Delta(\cC)$ the set of probability distributions on $\cC$ and $\mu(A) := \sum_{c \in A} \mu(c)$ for $A \subseteq \cC$), the cell-mass entropy is $H_W(\mu) := \sum_{\tau \in \cC|_W :\, \mu(C_\tau) > 0} \mu(C_\tau)\log_2\frac{1}{\mu(C_\tau)}$.

\paragraph{The singleton class and the uniform low-error distribution.}
Fix an error parameter $\delta \in (0, 1/4]$ and an integer $m > 1/\delta$. Let $\cX := \{x_1, \dots, x_m\}$ be a set of $m$ distinct points, and let $\cC := \{c_0, c_1, \dots, c_m\} \subseteq \{0,1\}^{\cX}$ consist of the all-zero concept and the $m$ singleton indicators (where $\ind[P]$ is $1$ if $P$ holds and $0$ otherwise):
\[
  c_0(x) := 0 \quad\text{for all } x \in \cX,
  \qquad
  c_i(x) := \ind[x = x_i] \quad\text{for } i \in \{1, \dots, m\}.
\]
The singleton $\{x_1\}$ is shattered by $\{c_0, c_1\}$, whereas no two-element set $\{x_i, x_j\}$ ($i \ne j$) is shattered because every $c \in \cC$ assigns label $1$ to at most one point of $\cX$; hence $\cC$ has VC dimension $\VC(\cC) = 1$. Consider the $\cC$-realizable sample $S := ((x_1, 0), \dots, (x_m, 0))$, realized by $c^* = c_0$. Since $c_i$ is the unique concept in $\cC$ with $c(x_i) = 1$, we have $\Pr_{c \sim \nu}[c(x_i) \ne 0] = \nu(c_i)$ for every $\nu \in \Delta(\cC)$ and $i \in \{1, \dots, m\}$. Thus at $t = 0$ (where $W = \emptyset$ and $\cP_0 = \{\cC\}$, so $\nu(\cC) = \mu_0(\cC) = 1$ holds automatically for every $\nu \in \Delta(\cC)$ and $(t + 1)\delta = \delta$), the feasible polytope is
\[
  \cK(\delta) \,:=\, \bigl\{\nu \in \Delta(\cC) : \forall i \in \{1, \dots, m\},\; \nu(c_i) \le \delta\bigr\}.
\]

Let $\mu \in \Delta(\cC)$ be the uniform distribution on $\{c_1, \dots, c_m\}$, so $\mu(c_0) = 0$ and $\mu(c_i) = 1/m$ for each $i \in \{1, \dots, m\}$. Because $m > 1/\delta$, we have $\mu(c_i) = 1/m < \delta$ for all $i$, so $\mu \in \cK(\delta)$ while being uniformly spread across $m$ distinct concepts (of Shannon entropy $\sum_{c \in \cC :\, \mu(c) > 0} \mu(c)\log_2\frac{1}{\mu(c)} = \log_2 m$, even though $H_\emptyset(\mu) = 0$). Yet for \emph{every} strict linear order $\prec$ on $\cC$ and \emph{every} subset $W \subseteq \cX$ of size $k := |W| < m$:
\begin{itemize}[leftmargin=1.8em, itemsep=2pt, topsep=3pt]
  \item Each $x_i \in W$ separates $c_i$ into its own singleton cell $\{c_i\}$ of mass $\mu(\{c_i\}) = 1/m$, while all remaining concepts in $\cC_0 := \cC \setminus \{c_i : x_i \in W\}$ satisfy $c(x) = 0$ for all $x \in W$ and stay together in a single cell of mass $\mu(\cC_0) = 1 - k/m > 0$.
  \item The representative $\min_\prec(\cC_0)$ of $\cC_0$ has mass at most $1/m$, so the total representative mass satisfies
  \[
    \mu(\Pool(W)) \,=\, \frac{k}{m} + \mu\bigl(\min_\prec(\cC_0)\bigr) \,\le\, \frac{k + 1}{m}.
  \]
  \item Applying $\ln u \le u - 1$ at $u = \frac{1}{1 - k/m} > 0$ gives $\ln\frac{1}{1 - k/m} \le \frac{k/m}{1 - k/m}$, so multiplying by $1 - k/m > 0$ and dividing by $\ln 2 > 1/2$ yields $\bigl(1 - \frac{k}{m}\bigr)\log_2\frac{1}{1 - k/m} \le \frac{k}{m \ln 2} \le \frac{2k}{m}$. Hence the cell-mass entropy satisfies
  \[
    H_W(\mu) \,=\, k \cdot \frac{1}{m}\log_2 m + \Bigl(1 - \frac{k}{m}\Bigr)\log_2\frac{1}{1 - k/m} \,\le\, \frac{k(\log_2 m + 2)}{m}.
  \]
\end{itemize}
For any size bound $k_0$ independent of the sample size $m$ (for instance, any function of $d$ and $\delta$ alone), once $m > k_0$ every subset $W \subseteq \cX$ with $|W| \le k_0$ satisfies $\mu(\Pool(W)) \le (k_0 + 1)/m$ and $H_W(\mu) \le k_0(\log_2 m + 2)/m$, both of which tend to $0$ as $m \to \infty$.

\paragraph{How the maximal lottery circumvents the obstruction.}
Suppose the order is $c_1 \prec c_2 \prec \dots \prec c_m \prec c_0$, placing the zero-error concept $c_0$ last. Under the maximal lottery game (Section~\ref{subsec:qem}), the uniform distribution on $\{c_1, \dots, c_m\}$ cannot be an equilibrium: moving mass from $c_m$ to $c_1$ strictly wins head-to-head and does not violate $x_1$'s error cap until $\mu(c_1)$ hits $\delta$. In equilibrium, $\mu^*$ is forced to pack mass greedily up to the $\delta$ error cap on the earliest singletons: $\mu^*(c_1) = \dots = \mu^*(c_{1/\delta}) = \delta$ (ignoring rounding), putting $0$ mass on the remaining singletons and on $c_0$. Now each of $c_1, \dots, c_{1/\delta}$ carries mass $\delta$ instead of $1/m$, so taking $W = \{x_1, \dots, x_{1/(2\delta)}\}$ ($O(1/\delta)$ points) splits off half of the total mass into singleton cells of mass $\delta$ and leaves the remaining cell with mass $1/2$; every cell then has mass at most $1/2$, giving $H_W(\mu^*) \ge 1$ bit of cell-mass entropy independent of $m$.

\end{document}